\documentclass{article}

\PassOptionsToPackage{numbers, compress}{natbib}
\usepackage[final]{neurips_2022}

\usepackage[utf8]{inputenc} % allow utf-8 input
\usepackage[T1]{fontenc}    % use 8-bit T1 fonts
\usepackage{hyperref}       % hyperlinks
\usepackage{url}            % simple URL typesetting
\usepackage{booktabs}       % professional-quality tables
\usepackage{amsfonts}       % blackboard math symbols
\usepackage{nicefrac}       % compact symbols for 1/2, etc.
\usepackage{microtype}      % microtypography
\usepackage{xcolor}         % colors
\usepackage{colortbl}
\usepackage{wrapfig}
\usepackage{graphicx}
\usepackage{lipsum}
\usepackage{multirow}
\usepackage{textcomp}
\usepackage{pifont} 
\newcommand{\cmark}{\ding{51}}%
\newcommand{\xmark}{\ding{55}}%

\usepackage{caption}
\usepackage{subcaption}

\title{Latency-aware Spatial-wise Dynamic Networks}

\newcommand{\nameShort}{LASNet}
\author{%
  Yizeng~Han$^1$\thanks{Equal contribution.}~~
  Zhihang~Yuan$^{2*}$~
  Yifan Pu$^{1*}$~
  Chenhao Xue$^2$\\ %\vspace{-6ex}\AND 
  \textbf{Shiji Song$^1$}~
  \textbf{Guangyu Sun$^2$}~
  \textbf{Gao Huang$^{1}$}\thanks{Corresponding author.}  \\ 
  $^1$ Department of Automation, BNRist, Tsinghua University, Beijing, China\\
  $^2$ School of Electronics Engineering and Computer Science, Peking University, Beijing, China\\
  \texttt{\{hanyz18, pyf20\}@mails.tsinghua.edu.cn}, \texttt{\{shijis, gaohuang\}@tsinghua.edu.cn} \\
  \texttt{\{yuanzhihang, xch927027, gsun\}@pku.edu.cn}\\
}

\begin{document}

\maketitle

%!TEX root = main.tex
\begin{abstract}
% \vskip -0.1in
    Spatial-wise dynamic convolution has become a promising approach to improving the inference efficiency of deep networks. By allocating more computation to the most informative pixels, such an \emph{adaptive} inference paradigm reduces the spatial redundancy in image features and saves a considerable amount of unnecessary computation. However, the \emph{theoretical} efficiency achieved by previous methods can hardly translate into a \emph{realistic} speedup, especially on the multi-core processors (\emph{e.g.} GPUs). The key challenge is that the existing literature has only focused on designing algorithms with minimal \emph{computation}, ignoring the fact that the practical latency can also be influenced by \emph{scheduling strategies} and \emph{hardware properties}. To bridge the gap between theoretical computation and practical efficiency, we propose a \emph{latency-aware} spatial-wise dynamic network (\nameShort), which performs \emph{coarse-grained} spatially adaptive inference under the guidance of a novel \emph{latency prediction model}. The latency prediction model can efficiently estimate the inference latency of dynamic networks by simultaneously considering algorithms, scheduling strategies, and hardware properties. We use the latency predictor to guide both the algorithm design and the scheduling optimization on various hardware platforms. Experiments on image classification, object detection and instance segmentation demonstrate that the proposed framework significantly improves the practical inference efficiency of deep networks. For example, the average latency of a ResNet-101 on the ImageNet validation set could be reduced by 36\% and 46\% on a server GPU (Nvidia Tesla-V100) and an edge device (Nvidia Jetson TX2 GPU) respectively without sacrificing the accuracy.
    Code is available at \url{https://github.com/LeapLabTHU/LASNet}.
  \end{abstract}

%!TEX root = main.tex

\section{Introduction}

\begin{figure}
\centering
\vskip -0.1in
\includegraphics[width=\textwidth]{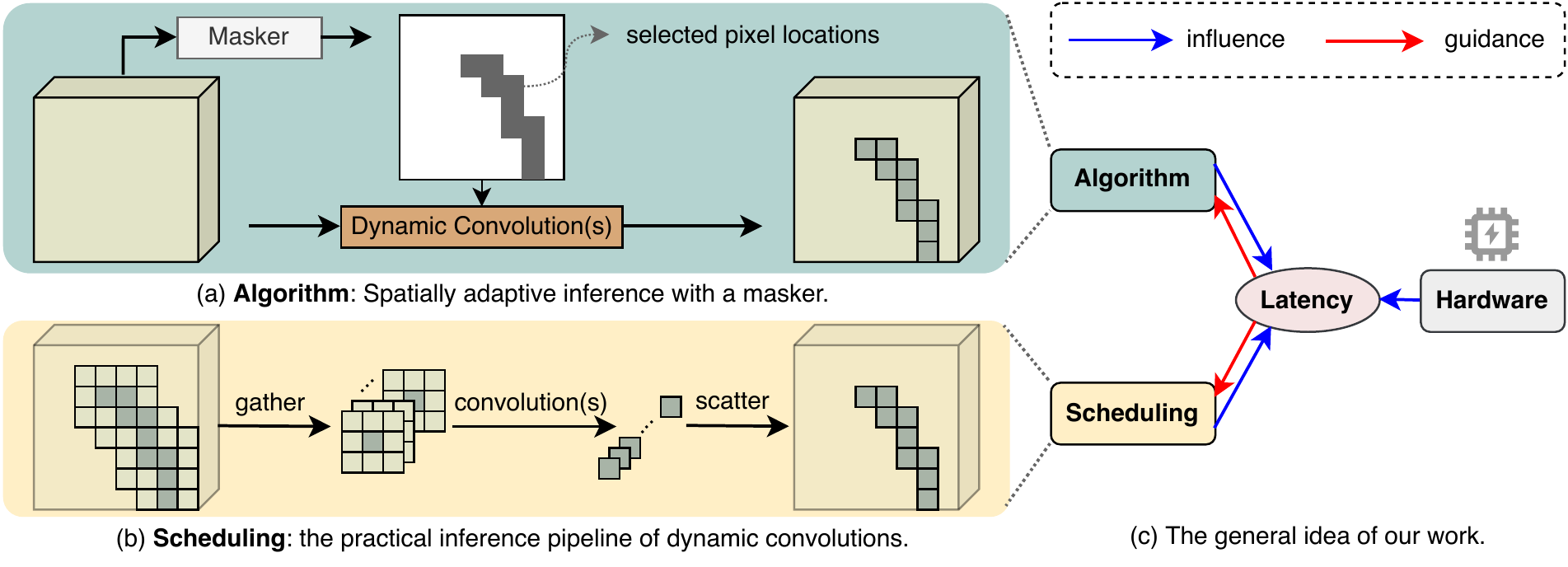}
\vskip -0.1in
\caption{An overview of our method. (a) illustrates the spatially adaptive inference \emph{algorithm}; (b) is the \emph{scheduling} strategy; and (c) presents the three key factors to the practical latency. For a given hardware, the latency is used to \emph{guide} our algorithm design and scheduling optimization.}
\vskip -0.2in
\label{fig:overview_dynconv}
\end{figure}

% Convolutional neural networks (CNNs) have achieved great success on many computer vision tasks~\cite{krizhevsky2012imagenet,he2016resnet,huang2017densely,lin2017feature,chen2017rethinking}. 
% Despite the remarkable performance of convolutional neural networks (CNNs) \cite{krizhevsky2012imagenet,he2016resnet,huang2017densely,lin2017feature,chen2017rethinking}, their high computational demand still hinders the employment and application of deep models on resource-constrained platforms. Recent work has focused on improving the inference efficiency of CNNs for better applicability. Popular approaches include designing lightweight network architectures~\cite{howard2017mobilenets, sandler2018mobilenetv2, ma2018shufflenet}, pruning redundant components~\cite{li2016pruning, he2018pruning} and performing adaptive inference \cite{huang2017multi,lin2017runtime, graves2016adaptive,han2021dynamic}. 
\vskip -0.1in
Dynamic neural networks \cite{han2021dynamic} have attracted great research interests in recent years. Compared to static models \cite{he2016resnet,huang2017densely,howard2017mobilenets,ma2018shufflenet} which treat different inputs equally during inference, {dynamic} networks can allocate the computation in a \emph{data-dependent} manner. For example, they can conditionally skip the computation of network layers \cite{huang2017multi,han2022learning,wang2018skipnet,veit2018convolutional} or convolutional channels \cite{lin2017runtime,bejnordi2019batch}, or perform \emph{spatially} adaptive inference on the most informative image regions (\emph{e.g.} the foreground areas) \cite{figurnov2017spatially,dong_more_2017,verelst_dynamic_2020,xie2020spatially,wang2020glance,SAR_TIP}. Spatial-wise dynamic networks, which typically decide whether to compute each feature pixel with \emph{masker} modules \cite{dong_more_2017,verelst_dynamic_2020,xie2020spatially,SAR_TIP} (\figurename~\ref{fig:overview_dynconv} (a)), have shown promising results in improving the inference efficiency of convolution neural networks (CNNs).
% Specifically, a lightweight module is usually used to generate a \emph{binary}-valued mask, which has the same spatial resolution as the output feature. Each pixel in the mask determines whether the corresponding feature pixel will be calculated or not (\figurename~\ref{}).

Despite the remarkable \emph{theoretical} efficiency achieved by spatial-wise dynamic networks \cite{dong_more_2017,verelst_dynamic_2020,xie2020spatially}, researchers have found it challenging to translate the theoretical results into \emph{realistic} speedup, especially on some multi-core processors, \emph{e.g.,} GPUs \cite{xie2020spatially,colleman2021processor,SAR_TIP}. The challenges are two-fold: 1) most previous approaches \cite{dong_more_2017,verelst_dynamic_2020,xie2020spatially} perform spatially adaptive inference at the finest \emph{granularity}: every \emph{pixel} is flexibly decided whether to be computed or not. Such flexibility induces non-contiguous \emph{memory access} \cite{xie2020spatially} and requires specialized \emph{scheduling strategies} (\figurename~\ref{fig:overview_dynconv} (b)); 2) the existing literature has only adopted the \emph{hardware-agnostic} FLOPs (floating-point operations) as an inaccurate proxy for the efficiency, lacking latency-aware guidance on the algorithm design. For dynamic networks, the adaptive computation with sub-optimal scheduling strategies further enlarges the discrepancy between the theoretical FLOPs and the practical latency. Note that it has been validated by previous works that the latency on CPUs has a strong correlation with FLOPs \cite{SAR_TIP,xie2020spatially}. Therefore, we mainly focus on the GPU platform in this paper, which is more challenging and less explored.

% its acceleration usually requires specialized scheduing strategies, while the current acceleration libraries (\emph{e.g.} cuDNN) are mostly developed for regular convolutions.
 
% \vskip -0.2in
% \lipsum[0.5]

We address the above challenges by proposing a \emph{latency-aware} spatial-wise dynamic network (\nameShort). Three key factors to the inference latency are considered: the \emph{algorithm}, the \emph{scheduling strategy}, and the \emph{hardware properties}. Given a target hardware device, we directly use the latency, rather than the FLOPs, to \emph{guide} our algorithm design and scheduling optimization (\figurename~\ref{fig:overview_dynconv} (c)).
% optimizing both the \emph{algorithm} (the granularity of spatially adaptive inference) and the \emph{scheduling strategies}, under the guidance of the \emph{inference latency} rather than the theoretical computation. \figurename~\ref{fig:overview_dynconv} (c) for an overview of our method.

% Designing latency-aware dynamic network algorithms is difficult, as it is laborious to measure the latency of dynamic operators on different hardware devices. 
Because the memory access pattern and the scheduling strategies in our dynamic operators differ from those in static networks, the libraries developed for static models (\emph{e.g.} cuDNN) are sub-optimal for the acceleration of dynamic models.
Without the support of libraries, each dynamic operator requires scheduling optimization, code optimization, compiling, and deployment for each device.
Therefore, it is laborious to evaluate the network latency on different hardware platforms.
To this end, we propose a novel \emph{latency prediction model} to efficiently estimate the realistic latency of a network by simultaneously considering the aforementioned three factors. Compared to the hardware-agnostic FLOPs, our predicted latency can better reflect the practical efficiency of dynamic models. 

Guided by this latency prediction model, we establish our latency-aware spatial-wise dynamic network (\nameShort), which adaptively decides whether to allocate computation on feature \emph{patches} instead of \emph{pixels} \cite{dong_more_2017,verelst_dynamic_2020,xie2020spatially} (\figurename~\ref{fig2_arch} top). We name this paradigm as spatially adaptive inference at a \emph{coarse granularity}. 
% Specifically, the decisions on whether to allocate computation are made for feature  
% For example, the feasible choices for the residual blocks in the first stage\footnote{Here we refer to stage as multiple network blocks, which process feature maps with the same resolution.} of a ResNet \cite{he2016resnet} include $\left\{1,2,4,7,8,14,28,56\right\}.$ 
While less flexible than the pixel-level adaptive computation in previous works \cite{dong_more_2017,verelst_dynamic_2020,xie2020spatially}, it facilitates more contiguous memory access, benefiting the realistic speedup on hardware. The scheduling strategy and the implementation are further ameliorated for faster inference. 
% A better trade-off between network performance and inference latency can be achieved by manipulating the granularity of spatially adaptive inference. 

% can trade off a certain flexibility for higher efficiency, to establish our hardware-friendly spatial-wise dynamic networks (\nameShort). 

% Second, 

% Second, we formulate the inference latency of spatial-wise dynamic networks as a function of \emph{algorithm (granularity), scheduling strategies} and \emph{hardware properties}. 
% we ameliorate the scheduling strategies for different granularities based on bottleneck structures in classic CNNs, \emph{e.g.,} RegNet \cite{radosavovic2020designing} and ResNet \cite{he2016resnet}. In basis of the optimized scheduling strategies, we model the relationship between latency and activation rate (the ratio of pixels that are calculated) of each block (\figurename~\ref{} for an example). It could be observed that there might be an optimal choice for different network stages. Inspired by this observation, we are able to establish \emph{Hardware-Friendly Spatial-wise Dynamic Networks} (\nameShort), which can achieve realistic speedup on GPU devices. 
% Moreover, by fitting the function of latency over activation rate in each block, the overall latency of processing each input image could be directly estimated. We further optimize the latency in the training phase to gain more significant improvements on the practical efficiency.

It is worth noting that \nameShort~is designed as a general framework in two aspects: 
1) the coarse-grained spatially adaptive inference paradigm can be conveniently implemented in various CNN backbones, \emph{e.g.,}  ResNets \cite{he2016resnet}, DenseNets \cite{huang2017densely} and RegNets \cite{radosavovic2020designing}; and
2) the latency predictor is an off-the-shell tool which can be directly used for various computing platforms (\emph{e.g.} server GPUs and edge devices).
% the relationship between latency and activation rate can also easily transfer among different network architectures; 
% 3) the overall formulation may also provide some insights for future works on the design of both static and dynamic networks.
% the latency-ware loss function can be directly used for training different spatial-wise dynamic models.

We evaluate the performance of \nameShort~on multiple CNN architectures on image classification, object detection, and instance segmentation tasks. Experiment results show that our \nameShort~improves the efficiency of deep CNNs both theoretically and practically. For example, the inference latency of ResNet-101 on ImageNet \cite{deng2009imagenet} is reduced by 36\% and 46\% on an Nvidia Tesla V100 GPU and an Nvidia Jetson TX2 GPU, respectively, without sacrificing the accuracy. Moreover, the proposed method outperforms various lightweight networks in a low-FLOPs regime.

Our main contributions are summarized as follows:
\begin{enumerate}

\item We propose \nameShort, which performs coarse-grained spatially adaptive inference guided by the practical latency instead of the theoretical FLOPs. To the best of our knowledge, \nameShort~is the first framework that directly considers the real latency in the design phase of dynamic neural networks;

\item We propose a latency prediction model, which can efficiently and accurately estimate the latency of dynamic operators by simultaneously considering the algorithm, the scheduling strategy, and the hardware properties;
    
\item Experiments on image classification and downstream tasks verify that our proposed \nameShort~can effectively improve the practical efficiency of different CNN architectures.
\end{enumerate}

%!TEX root = main.tex

% \vskip -0.2in
\section{Related works}
\vskip -0.1in
\noindent\textbf{Spatial-wise dynamic network} is a common type of dynamic neural networks \cite{han2021dynamic}. Compared to static models which treat different feature locations evenly during inference, these networks perform spatially adaptive inference on the most informative regions (\emph{e.g.}, foregrounds), and reduce the unnecessary computation on less important areas (\emph{e.g.}, backgrounds). Existing works mainly include three levels of dynamic computation: resolution level \cite{yang2020resolution,zhu2021dynamic}, region level \cite{wang2020glance} and pixel level \cite{dong_more_2017,verelst_dynamic_2020,xie2020spatially}. The former two generally manipulate the network inputs \cite{wang2020glance,zhu2021dynamic} or require special architecture design \cite{yang2020resolution}. In contrast, pixel-level dynamic networks can flexibly skip the convolutions on certain {feature} pixels in arbitrary CNN backbones \cite{dong_more_2017,verelst_dynamic_2020,xie2020spatially}. Despite its remarkable \emph{theoretical} efficiency, pixel-wise dynamic computation brings considerable difficulty to achieving \emph{realistic} speedup on multi-core processors, \emph{e.g.}, GPUs. Compared to the previous approaches \cite{dong_more_2017,verelst_dynamic_2020,xie2020spatially} which only focus on reducing the theoretical computation, we propose to directly use the latency to guide our algorithm design and scheduling optimization.

\noindent\textbf{Hardware-aware network design.} To bridge the gap between theoretical and practical efficiency of deep models, researchers have started to consider the real latency in the network design phase. There are two lines of works in this direction. One directly performs speed tests on targeted devices, and summarizes some guidelines to facilitate \emph{hand-designing} lightweight models \cite{ma2018shufflenet}. The other line of work \emph{searches} for fast models using the neural architecture search (NAS) technique \cite{tan2019mnasnet, wu2019fbnet}. However, all existing works try to build \emph{static} models, which have intrinsic computational redundancy by treating different inputs in the same way. However, speed tests for dynamic operators on different hardware devices can be very laborious and impractical. In contrast, our proposed latency prediction model can efficiently estimate the inference latency on any given computing platforms by simultaneously considering algorithm design, scheduling strategies and hardware properties.

% To the best of our knowledge, we are the first to study the realistic speedup of \emph{dynamic} neural networks by simultaneously considering algorithm design, scheduling strategies and hardware properties.
% combines the actual hardware latency with the neural architecture search to design the hardware-aware network architecture . However, these works rarely take dynamic networks into consideration.

% R

% HandCraft Design
% % ShuffleNetV2

% Hardware-aware NAS
% % MnasNet: Platform-Aware Neural Architecture Search for Mobile

% Sparse Conv Accelerate
% Scnn: An accelerator for compressed-sparse convolutional neural networks. ( designed a hardware  accelerator for sparse convolution and demonstrate that ideal speedup of sparse convolution is achievable in such devices.)
% Sbnet: Sparse blocks network for fast inference. ( a general method to accelerate sparse convolution in GPU)

% Processor Design
% Processor Architecture Optimization for Spatially Dynamic Neural Networks

% \noindent\textbf{Latency prediction and scheduling optimization: yzh} 

%!TEX root = main.tex

\section{Methodology}
\vskip -0.1in
In this section, we first introduce the preliminaries of spatially adaptive inference, and then demonstrate the architecture design of our \nameShort. The latency prediction model is then explained, which guides the granularity settings and the scheduling optimization for \nameShort. We further present the implementation improvements for faster inference, followed by the training strategies.

\subsection{Preliminaries}\label{sec_preliminary}
\vskip -0.1in
\noindent\textbf{Spatially adaptive inference.} The existing spatial-wise dynamic networks are usually established by attaching a masker $\mathcal{M}$ in each convolutional block of a CNN backbone (\figurename~\ref{fig:overview_dynconv} (a)). Specifically, let $\mathbf{x}\!\in\!\mathbb{R}^{H\times W\times C}$ denote the input of a block, where $H$ and $W$ are the feature height and width, and $C$ is the channel number. The masker $\mathcal{M}$ takes $\mathbf{x}$ as input, and generates a binary-valued spatial mask $\mathbf{M}\!=\!\mathcal{M}(\mathbf{x})\!\in\!\left\{0,1\right\}^{H\times W}$. Each element of $\mathbf{M}$ determines whether to perform convolution operations on the corresponding location of the output feature. The unselected regions will be filled with the values from the input \cite{dong_more_2017,verelst_dynamic_2020} or obtained via interpolation \cite{xie2020spatially}. We define the \emph{activation rate} of a block as $r\!=\!\frac{\sum_{i,j}\mathbf{M}_{i,j}}{H\times W}$, representing the ratio of the calculated pixels.

\noindent\textbf{Scheduling strategy.} 
% In the training stage, straight-through Gumbel Softmax \cite{jang2016categorical,maddison2016concrete} is a commonly used technique to optimize the non-differentiable mask generator $\mathcal{M}$ (\figurename~\ref{fig2_arch} top). One may optimize the activation rate (the ratio of selected/computed pixels) of each block \cite{xie2020spatially,SAR_TIP} or the overall FLOPs of the network \cite{verelst_dynamic_2020}.
During inference, the current scheduling strategy for spatial-wise dynamic convolutions generally involve three steps \cite{ren_sbnet_2018} (\figurename~\ref{fig:overview_dynconv} (b)): 1) \emph{gathering}, which re-organizes the selected pixels (if the convolution kernel size is greater than $1\times 1$, the neighbors are also required) along the \emph{batch} dimension; 2) \emph{computation}, which performs convolution on the gathered input; and 3) \emph{scattering}, which fills the computed pixels on their corresponding locations of the output feature. 
% Although this procedure reduces the \emph{computation}, it brings extra memory access, which might increase the practical latency.

\noindent\textbf{Limitations.} Compared to performing convolutions on the entire feature map, the aforementioned scheduling strategy reduces the computation while bringing considerable overhead to \emph{memory access} due to the mask generation and the non-contiguous memory access. Such overhead would increase the overall latency, especially when the \emph{granularity} of dynamic convolution is at the finest pixel level. 
% Moreover, the spatial size of the gathered pixels (patches) is relatively small, and the ``\emph{batch}'' size (\emph{i.e.} the number of selected pixels) is data-conditioned. Such flexibility also raise challenges on XXX. 

\begin{figure}[t]
    \begin{center}
    % \vskip -0.2in
      \includegraphics[width=\linewidth]{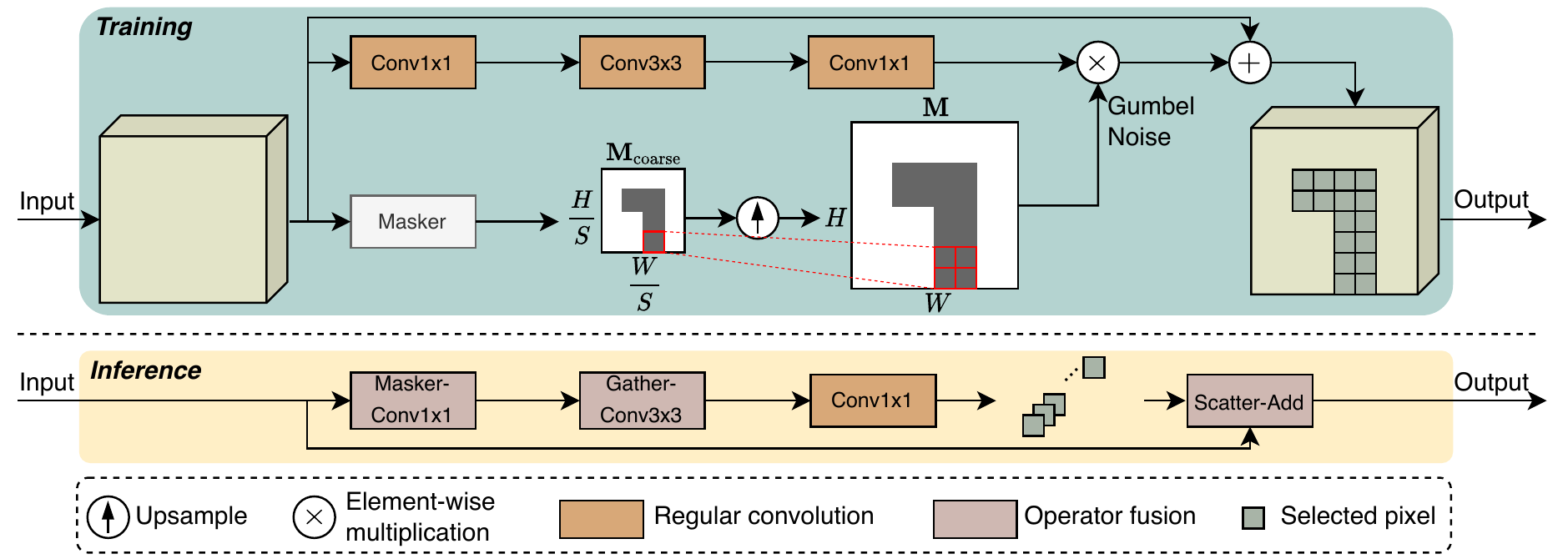}
    \end{center}
    \vskip -0.05in
    \caption{Our proposed \nameShort~block. Top: we first generate a low-resolution spatial mask $\mathbf{M}_{\mathrm{coarse}}$, which is then upsampled to obtain the mask $\mathbf{M}$ with the same size as the output feature. Gumbel Softmax \cite{jang2016categorical,maddison2016concrete} is used for end-to-end training (Sec.~\ref{sec_train}). Bottom: the scheduling optimization is performed to decrease the memory access for faster inference (Sec.~\ref{sec_schedule_optim}).}
    % following \cite{verelst_dynamic_2020}, we dilate the mask $\mathbf{M}$ to obtain the essential pixel locations for the $3\times 3$ convolution. To mitigate the extra burden on the memory access, we combine the gathering operation with the first $1\times 1$ convolution, and the scattering operation is implemented together with adding. See details in texts.}
    \vskip -0.2in
    \label{fig2_arch}
\end{figure}

\subsection{Architecture design}\label{sec_arch}
\vskip -0.1in
% As mentioned above, state-of-the-art spatial-wise dynamic CNNs typically use a masker module to decide whether to compute each \emph{pixel} on the output feature \cite{dong_more_2017,verelst_dynamic_2020,xie2020spatially}. Such flexibility brings substantial challenges to achieving realistic speedup on multi-core processors such as GPUs \cite{xie2020spatially,han2021dynamic}. 

\noindent\textbf{Spatial granularity.} As mentioned above, \emph{pixel}-level dynamic convolutions \cite{dong_more_2017,verelst_dynamic_2020,xie2020spatially} raise substantial challenges to achieving realistic speedup on multi-core processors due to the non-contiguous memory access. To this end, we propose to optimize the \emph{granularity} of spatially adaptive inference. Specifically, take the commonly used bottleneck structure in \cite{he2016resnet} as an example, our coarse-grained spatial-wise dynamic convolutional block is illustrated in \figurename~\ref{fig2_arch}. Instead of directly producing a mask with the shape of $H\!\times\! W$, we first generate a low-resolution mask $\mathbf{M}_{\mathrm{coarse}}\!\in\!\{0,1\}^{\frac{H}{S}\times\frac{W}{S}}$, where $S$ is named as the \emph{spatial granularity}. Each element in $\mathbf{M}_{\mathrm{coarse}}$ determines the computation of an $S\!\times \!S$-sized feature patch. For instance, the feature size in the first ResNet stage\footnote{Here we refer to a stage as the cascading of multiple blocks which process features with the same resolution.} is $56\times 56$. Then the possible choices for $S$ are $\left\{1,2,4,7,8,14,28,56\right\}.$ The mask $\mathbf{M}_{\mathrm{coarse}}$ is then upsampled to the size of $H\!\times\! W$. Notably, $S=1$ means that the granularity is still at the pixel level as previous methods \cite{dong_more_2017,verelst_dynamic_2020,xie2020spatially}. Note that the other extreme situation ($S=56$) is not considered in this paper, when the masker directly determines whether to skip the whole block (\emph{i.e.} layer skipping \cite{veit2018convolutional, wang2018skipnet}). Such an overly aggressive approach will lead to considerable drop of accuracy, as we presented in Appendix~\ref{supp_results_IN_cls}. The masker is composed of a pooling layer followed by a $1\times 1$ convolution. 
% While its theoretical computation is neglectable compared to the convolutions in the backbone, extra memory access will be caused by the masker module. We perform operator fusion to alleviate this issue (see details in Sec.~\ref{sec_schedule_optim}).

\noindent\textbf{Differences to existing works.} Without using the interpolation operation \cite{xie2020spatially} or the carefully designed two-branch structure \cite{SAR_TIP}, the proposed block architecture is simple and sufficiently general to be plugged into most backbones with minimal modification. Our formulation is mostly similar to that in \cite{verelst_dynamic_2020}, which could be viewed as a variant of our method with the spatial granularity $S$=1 for all blocks. Instead of performing spatially adaptive inference at the finest pixel level, our granularity $S$ is optimized under the guidance of our \emph{latency prediction model} (details are presented in the following Sec.~\ref{sec_latency_pred}) to achieve \emph{realistic speedup} on target computing platforms. 
% In contrast to \cite{SAR_TIP} which heuristically set the spatial granularity $S$, we propose to determine $S$ under the guidance of our \emph{latency prediction model} (details are presented in the following Sec.~\ref{sec_latency_pred}).

\subsection{Latency prediction model}\label{sec_latency_pred}
% \vskip -0.1in
% The inference latency on hardware is an important metric to evaluate the efficiency of deep networks.
% Because the memory access pattern and the scheduling strategies in our dynamic operators differ from those in static networks, the libraries of the static network (such as cuDNN) cannot be directly used for dynamic models.
% Without support of libraries, each dynamic operator requires scheduling optimization, code optimization, compiling, and deployment for each device and each layer.
% Therefore, it is time-consuming to evaluate the network performance on different hardware platforms.
% # Therefore, most of the current work uses extremely inaccurate theoretical calculations to estimate the delay
As stated before, it is laborious to evaluate the latency of dynamic operators on different hardware platforms. To efficiently seek preferable granularity settings on arbitrary hardware devices, we propose a latency prediction model $\mathcal{G}$, which can directly \emph{predict} the delay of executing dynamic operators on any target devices. 
% For a spatial-wise dynamic convolutional block, its \emph{activation rate} $r$ determines the computation. 
% As shown in Eq~\ref{latency_prediction_model_eq}, 
For a spatial-wise dynamic convolutional block, the latency predictor $\mathcal{G}$ takes the hardware properties $\mathbf{H}$, the layer parameters $\mathbf{P}$, the spatial granularity $S$, and the activation rate $r$ as input and predicts the latency $\ell$ of a dynamic convolutional block: $\ell=\mathcal{G}(\mathbf{H},\mathbf{P},S,r).$
% \begin{equation}
% \label{latency_prediction_model_eq}
% % \setlength{\abovedisplayskip}{-0.05ex}
%     \ell=\mathcal{G}(\mathbf{H},\mathbf{P},S,r).
%     \setlength{\belowdisplayskip}{-0.15ex}
% \end{equation}
% where $\ell$ is the predicted inference latency of the dynamic convolutional block.

\begin{figure} 
% \vskip -0.2in
  \begin{minipage}[t]{0.45\linewidth} 
    \centering 
    \includegraphics[width=\linewidth]{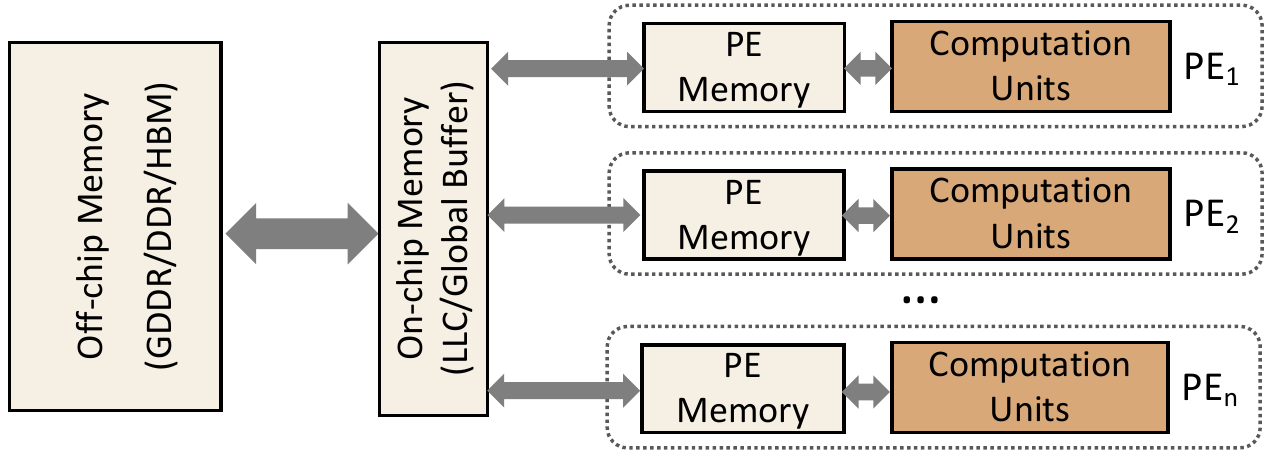} 
    \vskip -0.05in
    \caption{Our hardware model.} 
    \label{hardware_model} 
  \end{minipage}
  \begin{minipage}[t]{0.535\linewidth} 
    \centering 
    \includegraphics[width=\linewidth]{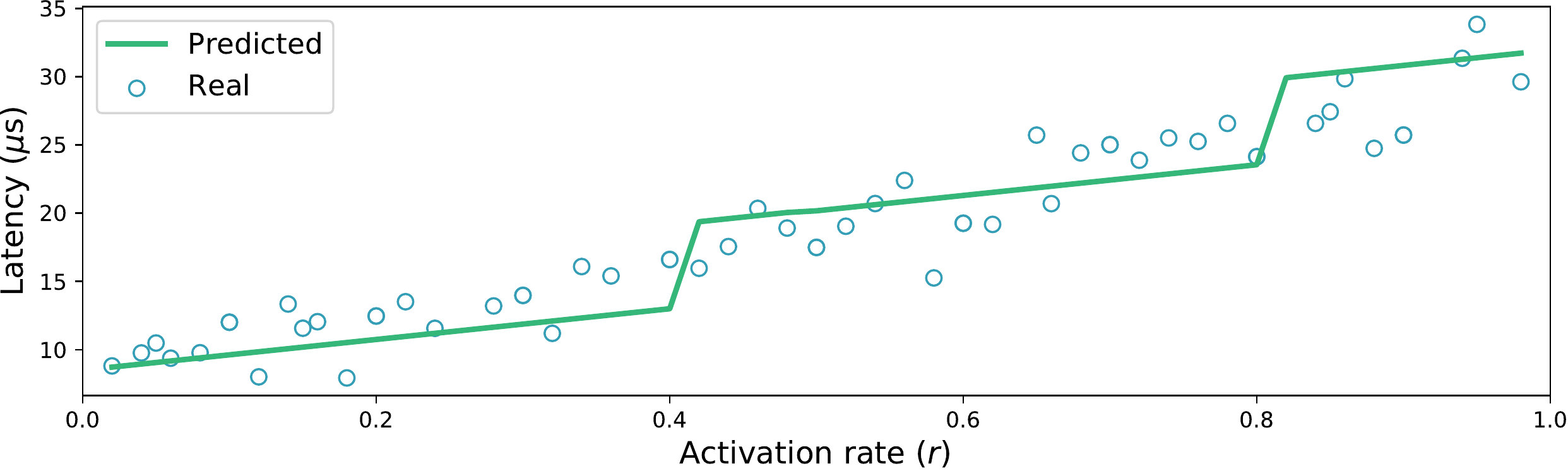} 
    \vskip -0.05in
    \caption{Latency prediction results.} 
    \label{real_predicted_latency} 
  \end{minipage} 
\vskip -0.25in
\end{figure}

\textbf{Hardware modeling.} We model a hardware device as multiple processing engines (PEs), and parallel computation can be executed on these PEs.
% For example, the processor engines of Nvidia GPU are streaming multiprocessors (SMs), which execute multiple threads in parallel.
% , and the processing engine of neural network accelerators executes the matrix multiplication with high parallelism. 
As shown in \figurename~\ref{hardware_model}, we model the memory system as a three-level structure \cite{hennessy2011computer}: 1) off-chip memory, 2) on-chip global memory, and 3) memory in PE. Such a hardware model enables us to accurately predict the cost on both \emph{data movement} and \emph{computation}.

\textbf{Latency prediction.} When simulating the {data movement} procedure, the efficiency of non-contiguous memory accesses under different granularity $S$ settings is considered. As for the computation latency, it is important to adopt a proper scheduling strategy to increase the parallelism of computation. Therefore, we search for the optimal scheduling (the configuration of tiling and in-PE parallel computing) of dynamic operations to maximize the utilization of hardware resources. A more detailed description of our latency prediction model is presented in Appendix~\ref{detailed_latency_predict}.
% the input/weight data from off-chip memory to each PE and moves the output data back to off-chip memory.
% It considers the efficiency of non-contiguous memory accesses. 
% due to dynamic inference.

% The on-chip memory is much faster than off-chip memory, but it has a limited size. Therefore, we cache the data on on-chip memory to improve performance.
% The computation of the dynamic operator should be distributed to different PEs.
% The predictor performs tiling on the output dimensions and assigns each tile to a PE.
% the computation of each output element to a PE by tiling of output channel dimension and the granularity dimension.
% Each tile is assigned to a PE.
% Because the number of active patches is unknown in the compilation stage, we process the patch dimension inside the PE to avoid the computing balance problem among different PEs.

\textbf{Empirical validation.} We take the first block in ResNet-101 as an example and vary the activation rate $r$ to evaluate the performance of our prediction model. The comparison between our predictions and the real testing latency on the Nvidia V100 GPU is illustrated in
\figurename~\ref{real_predicted_latency}, from which we can observe that our predictor can accurately estimate the real latency in a wide range of activation rates. 
% The small error is generally induced by 

% There are different memory units in various hardware devices, including off-chip DRAM and on-chip caches/buffers.

\subsection{Implementation details}\label{sec_schedule_optim}
% \noindent\textbf{Operator fusion}
\vskip -0.1in
We use general optimization methods like fusing activation functions and batch normalization layers into convolution layers.
We also optimize the specific operators in our spatial-wise dynamic convolutional blocks as follows (see also \figurename~\ref{fig2_arch} for an overview).

\textbf{Fusing the masker and the first convolution.}
As mentioned in Sec.~\ref{sec_preliminary}, the masker in each block consumes very little computation, but it takes the whole feature map as input.
Therefore, it is a \emph{memory-bounded} operation (the inference time is mainly spent on memory access).
Since the masker and the first convolution in the block share the same input, there is an opportunity to fuse these two operations to avoid the repeated access of the input data.
Note that a spatial-wise dynamic convolution requires the output of the masker.
If we fuse the two layers, the first convolution will be changed to a static operation, which may increase the inference latency.
% The smaller activation rate $r$, the larger the increase of the latency.
There exists a threshold of activation rate $r_{\mathrm{th}}$, when $r>r_{\mathrm{th}}$, the overall latency can be reduced.
We decide whether to fuse them according to the average activation rate. See more details in Appendix~\ref{sec_detailed_settings_for_latency_predict}.
% ~\footnote{Our experiments show most blocks should fuse the masker and the first convolution. }. 

\textbf{Fusing the gather operation and the dynamic convolution.}
Traditional approaches first gather the input pixels of the first dynamic convolution in a block (\figurename~\ref{fig:overview_dynconv} (b)).
% The gather operation extracts the activation patches from the input feature map.
The gather operation is also a \emph{memory-bounded} operation.
% It has a large amount of data input and output, and also has a large fetch pressure.
Furthermore, when the size of the convolution kernel exceeds 1$\times$1, the area of input patches may overlap, resulting in repeated memory load/store.
% the input patch size will expand.
% If there exists adjacent patches, 
We fuse the gather operation into the dynamic convolution to reduce the memory access.

\textbf{Fusing the scatter operation and the add operation.}
Traditional approaches scatter the output pixels of the last dynamic convolution, and then execute the element-wise addition (\figurename~\ref{fig:overview_dynconv} (b)).
We fuse these two operators to reduce the memory access. The ablation study in Sec.~\ref{sec_ablation} validates the effectiveness of the proposed fusing methods.

\subsection{Training}\label{sec_train}
\noindent\textbf{Optimization of non-differentiable maskers.} The masker modules are required to produce binary-valued spatial masks for making discrete decisions, and cannot be directly optimized with back propagation. 
Following \cite{xie2020spatially,verelst_dynamic_2020,SAR_TIP}, we adopt straight-through Gumbel Softmax \cite{jang2016categorical,maddison2016concrete} to train the network in an end-to-end fashion. 
% See details in the supplementary material.
Specifically, let $\tilde{\mathbf{M}}\!\in\!\mathbb{R}^{H\times W\times 2}$ denote the output of the mask generator. The decisions are obtained with the argmax function during inference. In the training phase, a differentiable approximation is defined by replacing the argmax operation with a Softmax:
\begin{equation}\small
    \hat{\mathbf{M}}=\frac{\exp\left\{\left(\log\left(\mathbf{\tilde{M}}_{:,:,0}\right)+\mathbf{G}_{:,:,0}\right)/\tau\right\}}{\sum_{k=0}^1 \exp\left\{\left(\log\left(\mathbf{\tilde{M}}_{:,:,k}\right)+\mathbf{G}_{:,:,k}\right)/\tau\right\}}\in[0,1]^{H\times W},
\end{equation}
where $\tau$ is the Softmax temperature. Following the common practice \cite{verelst_dynamic_2020,SAR_TIP}, we let $\tau$ decrease exponentially from 5.0 to 0.1 in training to facilitate the optimization of maskers. 

\noindent\textbf{Training objective.} 
% We define the \emph{activation rate} in a block as $r=\frac{\sum_{i,j}\mathbf{M}_{i,j}}{H\times W}$, representing the ratio of the calculated pixels. 
% The FLOPs of a block could be calculated based on the convolution configurations together with the activation rate $r$ \cite{xie2020spatially}. 
% Given the convolutional kernel size and the input/output feature sizes, 
The FLOPs of each spatial-wise dynamic convolutional block can be calculated based on our defined activation rate $r$ \cite{verelst_dynamic_2020}. Then we can obtain the FLOPs of the overall dynamic network $F_{\mathrm{dyn}}$. Let $F_{\mathrm{stat}}$ denotes the FLOPs of its static counterpart. We optimize their ratio to approximate a target $0<t<1$:$L_{\mathrm{FLOPs}}=(\frac{F_{\mathrm{dyn}}}{F_{\mathrm{stat}}}-t)^2.$
% \begin{equation}\label{eq_loss_flops}
%     L_{\mathrm{FLOPs}}=(\frac{F_{\mathrm{dyn}}}{F_{\mathrm{stat}}}-t)^2, 0<t<1.
% \end{equation}
In addition, we define loss item $L_{\mathrm{bounds}}$ as in \cite{verelst_dynamic_2020} to constrain the upper bound and the lower bound of activation rates in early training epochs.

% As mentioned in Sec.~\ref{sec_preliminary}, existing methods train spatial-wise dynamic networks to achieve a better trade-off between performance and the \emph{theoretical} efficiency: $L=L_{\mathrm{cls}}+\alpha L_\mathrm{act}$ \cite{xie2020spatially,SAR_TIP} or $L=L_{\mathrm{cls}}+\alpha L_\mathrm{FLOPs}$ \cite{verelst_dynamic_2020}, where $L_\mathrm{cls}$ is the cross-entropy loss for classification. The loss item $L_\mathrm{act}$ is to encourage the activation rates of different blocks to approximate some given targets, and $L_\mathrm{FLOPs}$ envolves the average FLOPs of the overall network. 

% In contrast, we can directly seek for an optimal trade-off between network performance and inference \emph{latency}, given that the relationship between latency and activation rate is obtained (Sec. \ref{sec_schedule_optim}):
% \begin{equation}
%     L = L_{\mathrm{cls}} + \alpha L_\mathrm{latency},
% \end{equation}
% where $L_\mathrm{latency}=XXX$.

 We further propose to leverage the static counterparts of our dynamic networks as ``teachers'' to guide the optimization procedure. Let $\mathbf{y}$ and $\mathbf{y}'$ denote the output logits of a dynamic model (``student'') and its ``teacher'', respectively. Our final loss can be written as
 \begin{equation}\label{eq_loss}
    L = L_{\mathrm{task}} + \alpha (L_\mathrm{FLOPs}+L_{\mathrm{bounds}}) + \beta T^2\cdot \mathrm{KL}(\sigma(\mathbf{y}/T)||\sigma(\mathbf{y}'/T)),
\end{equation}
where $L_\mathrm{task}$ represents the task-related loss, \emph{e.g.}, cross-entropy loss in image classification. $\mathrm{KL}(\cdot||\cdot)$ denotes the Kullback–Leibler divergence, and $\alpha,\beta$ are the coefficients balancing these items. We use $\sigma$ to denote the log-Softmax function, and $T$ is the temperature for computing KL-divergence.
%  See Sec. \ref{sec_ablation} for our ablation studies.

%!TEX root = main.tex

\section{Experiments}
\vskip -0.1in
In this section, we first introduce the experiment settings in Sec.~\ref{sec_setup}. Then the latency of different granularity settings are analyzed in Sec.~\ref{sec_latency_pred}. The performance of our \nameShort~on ImageNet is further evaluated in Sec.~\ref{sec_IN_results}, followed by the ablation studies in Sec.~\ref{sec_ablation}. Visualization results are illustrated in Sec.~\ref{sec_vis}, and we finally validate our method on the object detection task (Sec.~\ref{sec_det}). The results on the instance segmentation task are presented in . For simplicity, we add ``LAS-'' as a prefix before model names to denote our \nameShort, \emph{e.g.}, LAS-ResNet-50.

% including both 
% We first present the results of the theoretical latency model, and compare the latency of each block under different granularities. Next, the accuracy and theoretical latency of ours designed model is presented. The results of training \nameShort with the latency-aware training objective will then be shown. Finally, we verify the effectiveness of ours method on downstream tasks, \emph{i.e.} object detection, semantic segmentation and human pose estimation.

\subsection{Experiment settings}\label{sec_setup}
% \vskip -0.1in
\noindent\textbf{Latency prediction.} Various types of hardware platforms are tested, including a server GPU (Tesla V100), a desktop GPU (GTX1080) and edge devices (\emph{e.g.}, Nvidia Nano and Jetson TX2). The major properties considered by our latency prediction model include the number of processing engines (\#PE), the floating-point computation in a processing engine (\#FP32), the frequency and the bandwidth. It can be observed that the server GPUs generally have a larger \#PE than the IoT devices. The batch size is set as 1 for all dynamic models and computing platforms.
% Table~\ref{tab_hardware_property}. It can be observed that the server-end V100 has a significantly larger number of processing engines (\#PE) than the IoT divice TX2. 
% The information of other types of computing devices (\emph{e.g.}, Nvidia RTX3090) is presented in the supplementaty material. 

% \begin{wraptable}{r}{7.5cm}
% \vskip -0.2in
% % \begin{table}
%   \caption{Hardware properties.}
%   \vskip -0.1in
%   \label{tab_hardware_property}
%   \centering
%   \begin{tabular}{ccccc}
%     \toprule
    
%     Name & \#PE     &    \#FP32  &   frequency   & bandwidth \\
%     \midrule
%     V100    &80 & 64 & 1500M & 700G \\
%     TX2 & 2 & 128 & 1300M & 59.7G \\
%     \bottomrule
%   \end{tabular}
% % \end{table}
% \vskip -0.1in
% \end{wraptable}

\begin{figure}[t]
     \centering
    %  \vskip -0.2in
     \begin{subfigure}[b]{\textwidth}
         \centering
         \includegraphics[width=\textwidth]{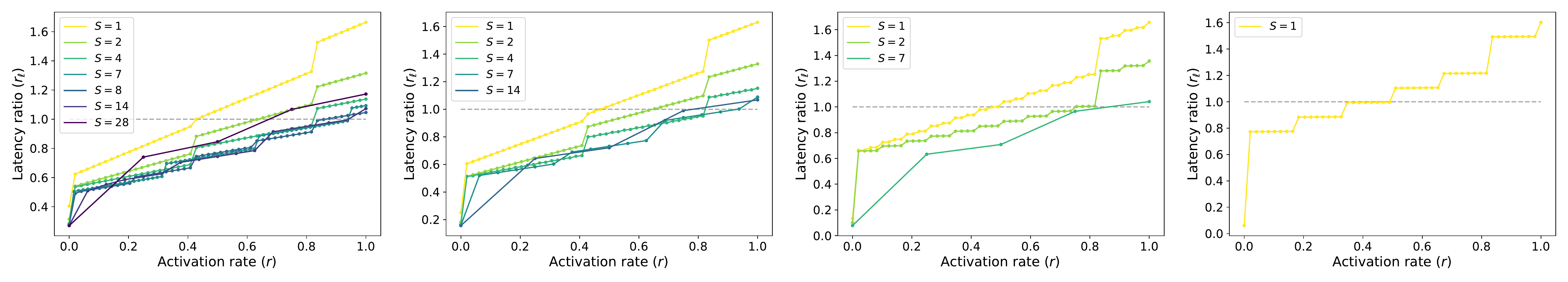}
         \vskip -0.1in
         \caption{Relationship between $r_{\ell}$ and $r$ for LAS-ResNet blocks on the Nvidia Tesla V100 GPU.}
         \label{fig_latency_pred_res_v100}
     \end{subfigure}
    %  \hfill
     \begin{subfigure}[b]{\textwidth}
         \centering
         \includegraphics[width=\textwidth]{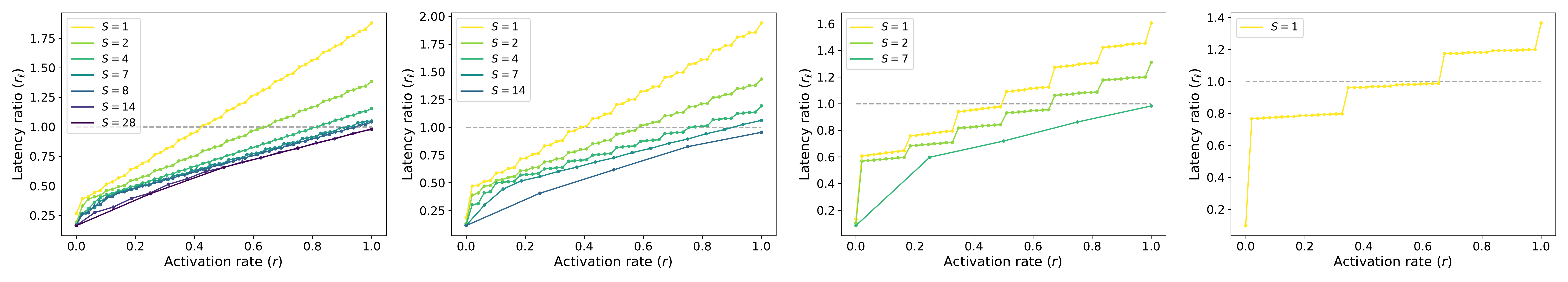}
         \vskip -0.1in
         \caption{Relationship between $r_{\ell}$ and $r$ for LAS-RegNetY-800MF blocks on the Nvidia Jetson TX2 GPU.}
         \label{fig_latency_pred_reg_tx2}
     \end{subfigure}
     \vskip -0.05in
        \caption{Latency prediction results of LAS-ResNet blocks on V100 (a) and LAS-RegNet blocks on TX2 (b). For both networks, we plot the relationship between the latency ratio $r_{\ell}$ and the activation rate $r$ for the blocks in 4 stages with the convolutional stride 1. 
        The practical efficiency is only improved when $r_{\ell}<1$. Note that $S=1$ can harm the practical latency even for a small $r$ (reduced computation), while a larger $S$ will alleviate this problem.
        See detailed analysis in Sec.~\ref{sec_latency_pred}.}
        \vskip -0.25in
        \label{fig_latency_pred}
\end{figure}

\noindent\textbf{Image classification.}
The image classification experiments are conducted on the ImageNet \cite{deng2009imagenet} dataset. 
% The ImageNet dataset \cite{deng2009imagenet} contains 1.2 million training images and 50,000 vaildation images in 1000 classes. 
Following \cite{verelst_dynamic_2020}, we initialize the backbone parameter from a pre-trained checkpoint\footnote{We use the torchvision pre-trained models at \url{https://pytorch.org/vision/stable/models.html}.}, and finetune the whole network for 100 epochs with the loss function in Eq.~(\ref{eq_loss}). We fix $\alpha=10,\beta=0.5$ and $T=4.0$ for all dynamic models. More details are provided in Appendix~\ref{sec_detailed_settings}.
% Each model is trained  with the batch size dependent on the model size and the GPU memory. 
% The initial learning rate is set as 0.02$\times$batch size/256, decaying with a cosine shape. 
% The Gumbel temperature is gradually annealed from 5 to 0.1.

\vskip -0.1in
\subsection{Latency prediction results}\label{sec_latency_pred}
% \vskip -0.1in
In this subsection, we present the latency prediction results of the spatial-wise dynamic convolutional blocks in two different models: LAS-ResNet-101 \cite{he2016resnet} (on V100) and LAS-RegNetY-800MF \cite{radosavovic2020designing} (on TX2). All the blocks have the bottleneck structure with different channel numbers and convolution groups, and the RegNetY is equipped with Squeeze-and-Excitation (SE) \cite{hu2018squeeze} modules.

We first define $\ell_{\mathrm{dyn}}$ as the latency of a spatial-wise dynamic convolutional block, and $\ell_{\mathrm{stat}}$ as that of a static block without a masker. Their ratio is denoted as $r_{\ell}=\frac{\ell_{\mathrm{dyn}}}{\ell_{\mathrm{stat}}}$. We investigate the relationship between $r_{\ell}$ and the activation rate $r$ (cf. Sec.~\ref{sec_train}) for different \emph{granularity} settings. The results in \figurename~\ref{fig_latency_pred} demonstrate that: 1) even equipped with our special optimization on the scheduling strategies, pixel-level spatially adaptive inference ($S$=1) \emph{cannot always} improve the practical efficiency. Such fine-grained adaptive inference is adopted by most previous works \cite{verelst_dynamic_2020,xie2020spatially}, and our result can explain the reason why they can only achieve realistic speedup on less powerful CPUs \cite{xie2020spatially} or specialized devices \cite{colleman2021processor}; 2) a proper granularity $S>1$ effectively alleviates this problem on both hardware devices. By setting $S>1$, realistic speedup could be achieved with larger activation rates.
% 3) the advantage of coarse-grained spatially adaptive inference ($S>1$) is more significant on the server GPU with a larger \#PE, as the finest granularity ($S=1$) brings considerable overhead on the memory access and is not friendly to parallel computation.
% 4) the relationship between $r_{\ell}$ and $r$ is piece-wise linear, as we observe that the fusion of our masker and the first 1$\times$1 convolution can be sub-optimal for some activation ratios. 
% Therefore, we decide whether to perform this operator fusion dependent on $r$, leading to the piece-wise linear function of $r_{\ell}$ over $r$.
% This is because ... (masker-conv1, dependent on $r$) (I cannot organize the language now, too tired).

The latency prediction results are further used to seek for a preferable granularity setting for the first 3 stages (we fix $S=1$ for the last stage, where the feature resolution is $7\times 7$). Therefore, we plot the relationship between $r_{\ell}$ and $S$ in \figurename~\ref{fig_latency_r_s}. It can be observed that: 1) $r_{\ell}$ generally decreases with $S$ increasing for a given $r$; 2) an overly large $S$ (less flexible adaptive inference) brings insignificant improvement on both devices. Especially, enlarging $S$ from 8 to 28 in the first stage of a LAS-ResNet brings very little improvement on V100. Based on the results in \figurename~\ref{fig_latency_r_s}, we can trade off between flexibility and efficiency by selecting appropriate $S$ for different models and hardware devices. For example, we can simply set $S_{\mathrm{net}}$=8-4-7-1\footnote{We use this form to represnet the $S$ settings for the 4 stages of a network.} in a LAS-ResNet-101 to achive realistic speedup. The accuracy-latency plots in \figurename~\ref{fig_main_results} also validate this observation.
More results of our latency prediction model on the desktop-level GPU, GTX1080, are presented in Appendix~\ref{supp_results_latency_pred}.

% V100, and even harms the practical efficiency on the less powerful TX2. 
% Therefore, 
% , including the latency of LAS-ResNet blocks on TX2 and the latency of LAS-RegNet blocks on V100.
% In this subsection, we illustrate the optimized scheduling for spatial-wise dynamic networks. 

% From figXX we can observe that, the most fine-grained spatial granularity (sg=1) is not the best in the sense of speed. This is because the operation need to start a separate thread for each patch. Use one pixel as a patch would let the system start too many threads thus drag the overall inference latency. Meanwhile, one thread is not enough to process a large patch, so too large patch will also drag back the inference speed.

\subsection{ImageNet classification results}\label{sec_IN_results}
\vskip -0.1in
% We select the faster granularity for each stage in RegNet and train the model. The accuracy-latency and accuracy-FLOPs results shows that although scale up granularity introduce no extra computation, the latency is reduce and the model performance also improves slightly.
We now empirically evaluate our proposed \nameShort~on the ImageNet dataset. The network performance is measured in terms of the trade-off between classification accuracy and inference efficiency. Both theoretical (\emph{i.e.} FLOPs) and practical efficiency (\emph{i.e.} latency) are tested in our experiments.
\begin{figure}
% \vskip -0.2in
     \centering
     \begin{subfigure}[b]{\textwidth}
         \centering
         \includegraphics[width=\textwidth]{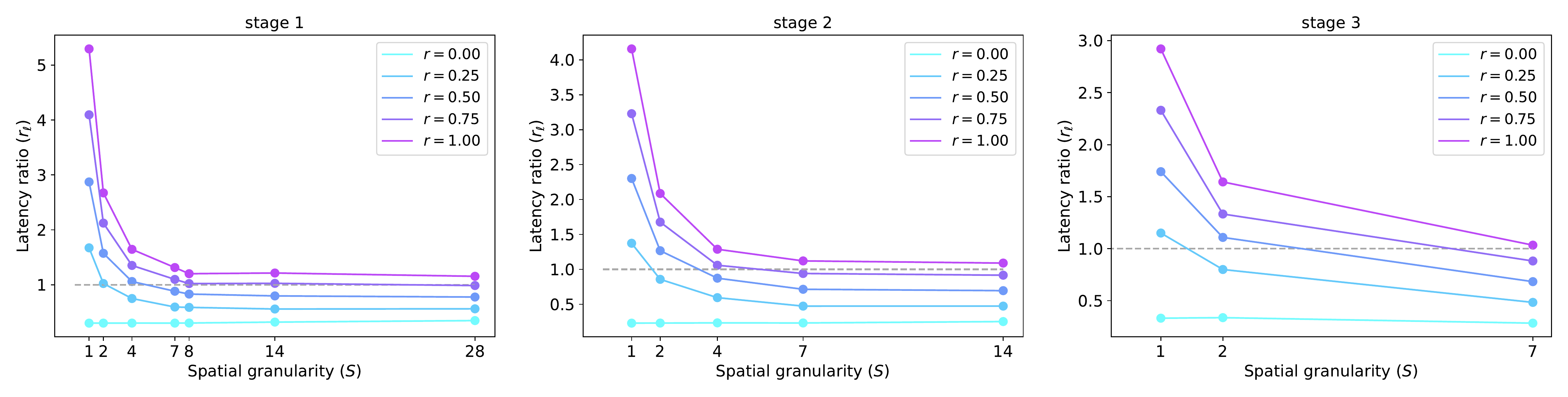}
         \vskip -0.1in
         \caption{Relationship between $r_{\ell}$ and $S$ for LAS-ResNet blocks on V100.}
         \label{fig_r_S_v100}
     \end{subfigure}
    %  \hfill
     \begin{subfigure}[b]{\textwidth}
         \centering
         \includegraphics[width=\textwidth]{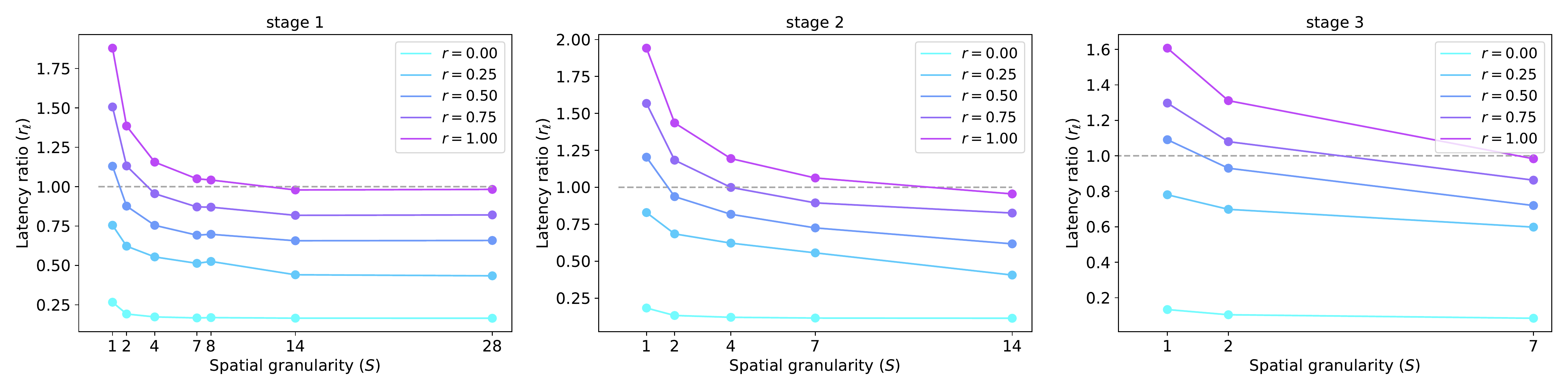}
         \vskip -0.1in
         \caption{Relationship between $r_{\ell}$ and $S$ for LAS-RegNetY-800MF blocks on Nvidia Jetson TX2 GPU.}
         \label{fig_r_S_tx2}
     \end{subfigure}
     \vskip -0.05in
        \caption{The relationship between the latency ratio $r_{\ell}$ and the spatial granularity $S$.}
        % Best viewed in color.}
        % For different activation rates, coarse-grained spatially adaptive inference ($S>1$) can significantly reduce the latency ratio $r_{\ell}$. }
         \vskip -0.2in
        \label{fig_latency_r_s}
\end{figure}

\subsubsection{Standard baseline comparison: ResNets}
\vskip -0.1in
We first establish our \nameShort~based on the standard ResNets \cite{he2016resnet}. Specifically, we build LAS-ResNet-50 and LAS-ResNet-101 by plugging our maskers in the two common ResNet structures.

\noindent\textbf{The baselines} include various types of dynamic inference approaches: 1) layer skipping (SkipNet \cite{wang2018skipnet} and Conv-AIG \cite{veit2018convolutional}); 2) channel skipping (BAS \cite{bejnordi2019batch}); and 3) pixel-level spatial-wise dynamic network (DynConv \cite{verelst_dynamic_2020}). For our \nameShort, we compare various settings of the spatial granularity $S_{\mathrm{net}}$. We set training targets (cf. Sec.~\ref{sec_train}) $t\!\in\!\{0,4,0.5,0.6,0.7\}$ for our dynamic models to evaluate their performance in different sparsity regimes. We apply the same operator fusion (Sec.~\ref{sec_schedule_optim}) for both our models and the compared baselines \cite{veit2018convolutional,verelst_dynamic_2020} for fair comparison.

\begin{figure}
     \centering
     \vskip -0.2in
     \begin{subfigure}[b]{\textwidth}
         \centering
         \includegraphics[width=\textwidth]{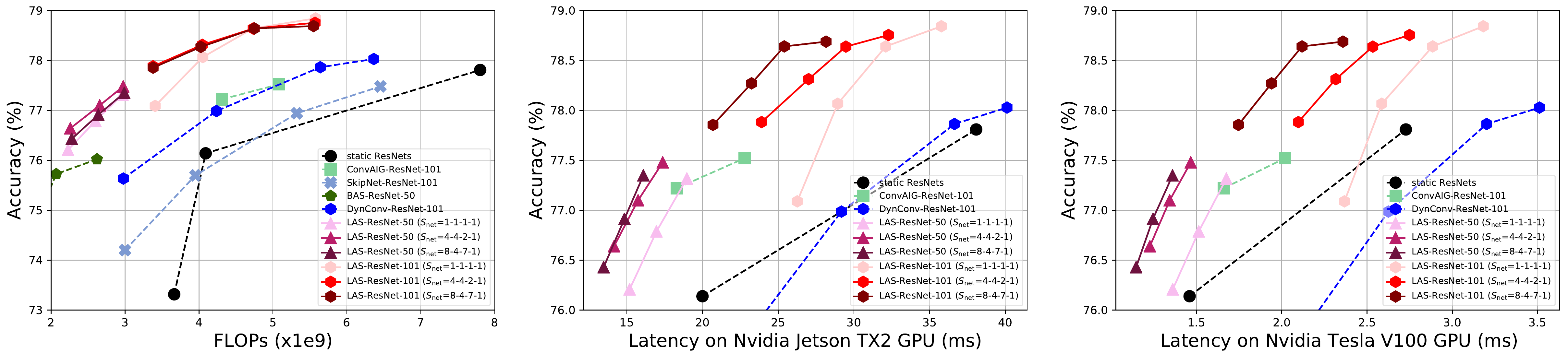}
         \vskip -0.05in
         \caption{LAS-ResNet results.}
         \label{fig_resnets}
     \end{subfigure}
    %  \hfill
     \begin{subfigure}[b]{\textwidth}
         \centering
         \includegraphics[width=\textwidth]{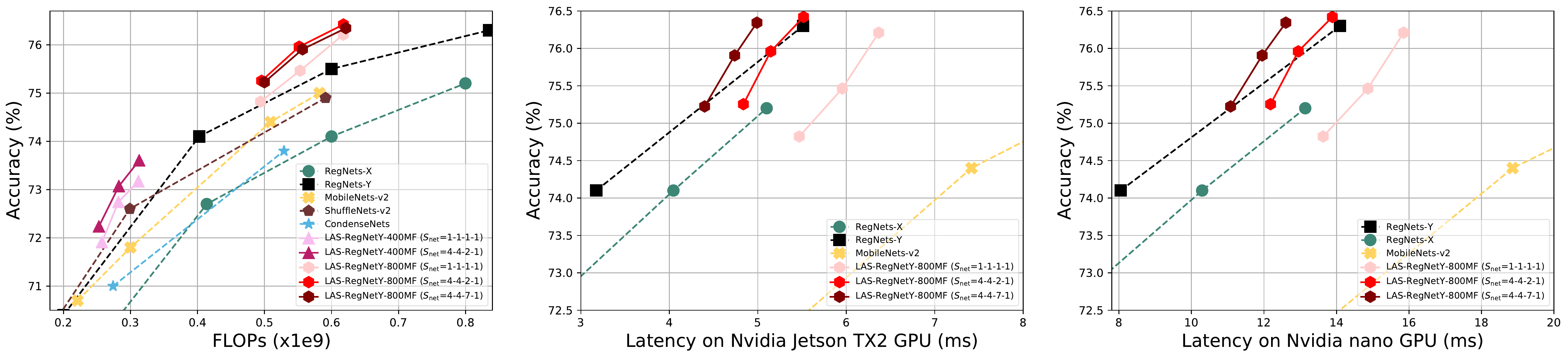}
         \vskip -0.05in
         \caption{LAS-RegNetY results.}
         \label{fig_regnets}
     \end{subfigure}
     \vskip -0.05in
        \caption{Experimental results on ImageNet. The proposed coarse-grained spatially adaptive inference is tested on standard ResNets (a) and lightweight RegNets (b).}
         \vskip -0.2in
        \label{fig_main_results}
\end{figure}

\noindent\textbf{Results} are presented in \figurename~\ref{fig_main_results} (a). On the left we plot the relationship of accuracy \emph{v.s.} FLOPs. It can be observed that our LAS-ResNets with different granularity settings significantly outperform the competing dynamic neural networks. Surprisingly, coarse-grained spatially adaptive inference ($S_{\mathrm{net}}$=4-4-2-1 and $S_{\mathrm{net}}$=8-4-7-1 for the 4 stages) can achieve even higher accuracy when consuming similar FLOPs on ResNets, despite the sacrificed flexibility compared to $S_{\mathrm{net}}$=1-1-1-1. We conjecture that a larger $S$ is also beneficial to the optimization of maskers.

We compare the practical latency of three granularity settings in \figurename~\ref{fig_main_results} (a) predicted by our latency prediction model (middle on TX2 and right on V100). We can witness that although they achieve comparable theoretical efficiency (\figurename~\ref{fig_main_results} (a) left), larger $S$ is more hardware-friendly compared to the finest granularity. For example, the inference latency of LAS-ResNet-101 ($S_{\mathrm{net}}$=1-1-1-1) is significantly higher than the ResNet-101 baseline on V100 (\figurename~\ref{fig_main_results} (a) right), even though its theoretical computation is much smaller than that of the static model. However, larger granularities ($S_{\mathrm{net}}$=4-4-2-1 and $S_{\mathrm{net}}$=8-4-7-1) can effectively improve the inference latency due to its lower burden on the memory access. 
% The superiority of coarse-grained spatially adaptive inference diminishes (but still exists) on the less powerful IoT device, Nvidia Jetson TX2 (\figurename~\ref{fig_resnets} right). The advantage is more significant on the Tesla V100 GPU, because of its large number of processing engines (\#PE). 
% Interestingly, the optimal settings for the two different hardware platforms are different ($S_{\mathrm{net}}$=4-4-2-1 is superior on TX2, while $S_{\mathrm{net}}$=8-4-7-1 leads to higher efficiency on V100). 
Remarkably, the latency of ResNet-101 could be reduced by 36\% and 46\% on V100 and TX2 respectively without sacrificing the accuracy when $t$=0.4. The classification accuracy is increased by 1.9\% with similar inference efficiency. It can be observed that the realistic speedup ratio $r_{\ell}$ is more close to the theoretical FLOPs ratio target $t$ on the less powerful TX2, because the latency is \emph{computation-bounded} (\emph{i.e.} the latency is mainly spent on computation) on such IoT devices. In contrast, there is a larger gap between practical and theoretical efficiency on the more powerful V100, as the latency is bounded by the memory access cost.

\subsubsection{Lightweight baseline comparison: RegNets}
% \vskip -0.1in
We further evaluate our \nameShort~in lightweight CNN architectures, \emph{i.e.} RegNets-Y \cite{radosavovic2020designing}. Two different sized models are tested: RegNetY-400MF and RegNetY-800MF. Compared baselines include other types of efficient models, \emph{e.g.}, MobileNets-v2 \cite{sandler2018mobilenetv2}, ShuffletNets-v2 \cite{ma2018shufflenet} and CondenseNets \cite{huang2018condensenet}.

The results are presented in \figurename~\ref{fig_main_results} (b). The x-axis for the three sub-figures are the FLOPs, the latency on TX2, and the latency on the Nvidia nano GPU, respectively. We can observe that our method outperforms various types of static models in terms of the trade-off between accuracy and efficiency. More results on image classification are provided in Appendix~\ref{supp_results_IN_cls}.

% \subsection{Latency-aware training objective results}

% We then replace the FLOPs penalty term in the DynConv loss function with latency penalty term in the training process. This would let the learning algorithm automatically find the fast model architecture. The results is show in figXX.

\subsection{Ablation studies}\label{sec_ablation}
% \vskip -0.1in
We conduct ablation studies to validate the effectiveness of our \emph{coarse-grained} spatially adaptive inference (Sec.~\ref{sec_arch}) and operator fusion operations (Sec.~\ref{sec_schedule_optim}).

\begin{figure}
% \vskip -0.1in
\centering
\begin{minipage}[t]{0.385\linewidth}
\centering
\includegraphics[width=\linewidth]{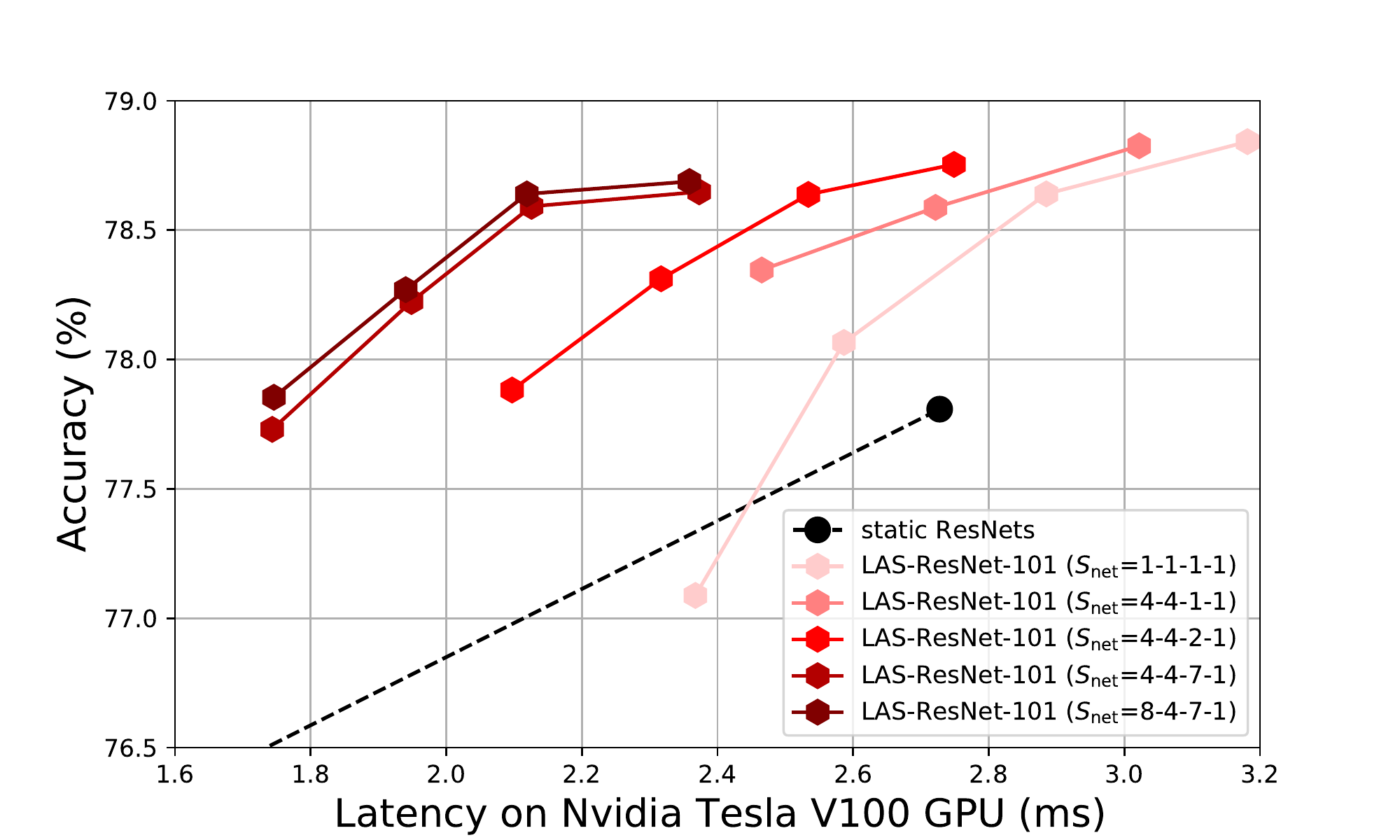}
\caption{Ablation studies on $S$.}
\label{fig_ablation_sg}
\end{minipage}
\begin{minipage}[t]{0.605\linewidth}
\centering
\includegraphics[width=\linewidth]{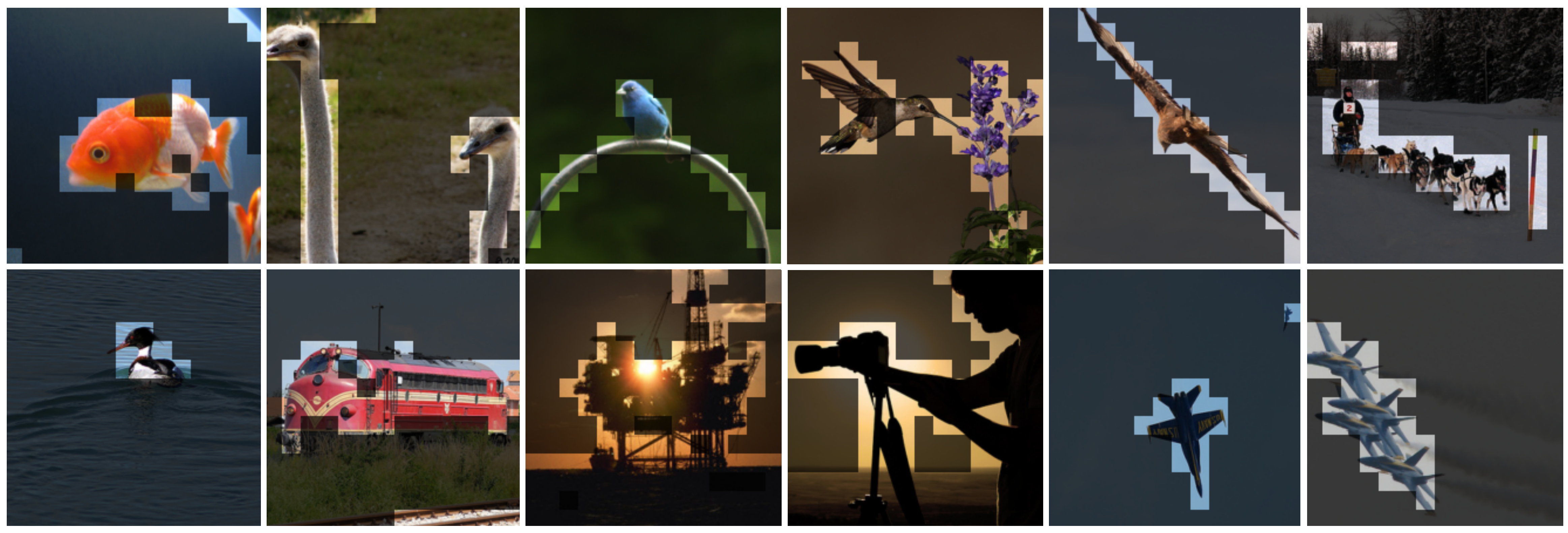}
\caption{Visualization results.}
\label{fig_vis}
\end{minipage}
\vskip -0.25in
\end{figure}

\noindent\textbf{More granularities settings.} We test various granularity settings on LAS-ResNet-101 to examine the effects of $S$ in different stages. The results on the Tesla-V100 GPU are presented in \figurename~\ref{fig_ablation_sg}. It can be found that the finest granularity ($S_{\mathrm{net}}$=1-1-1-1) leads to substantial inefficiency despite the reduced FLOPs (cf. \figurename~\ref{fig_main_results} (a) left). Coarse-grained spatially adaptive inference in the first two stages ($S_{\mathrm{net}}$=4-4-1-1) effectively reduces the inference latency. We further increase $S$ in the third stage to 2 and 7, and this procedure consistently improves the realistic efficiency on the V100 GPU. This trend also holds on the GTX 1080 GPU (see the results in Appendix~\ref{supp_results_IN_cls}.

\begin{wraptable}{r}{6.5cm}
% \begin{table}
\vskip -0.18in
  \caption{Ablation studies on operator fusion.}
  \vskip -0.05in
  \label{ablation_op_fusion}
%   \centering
  \begin{tabular}{cccc}
    \toprule
    Masker- & Gather- & Scatter-     & \multirow{1}*{Latency} \\
    Conv1x1 & Conv3x3 & Add & (\textmu s) \\
    \midrule
    \xmark & \xmark & \xmark & 163.2 \\
    \cmark & \xmark & \xmark & 90.1 \\
    \cmark & \cmark & \xmark & 86.7 \\
    \cellcolor{lightgray!50}\cmark & \cellcolor{lightgray!50}\cmark & \cellcolor{lightgray!50}\cmark & \cellcolor{lightgray!50}\textbf{71.4} \\
    \bottomrule
  \end{tabular}
  \vskip -0.1in
% \end{table}
\end{wraptable}

\noindent\textbf{Operator fusion.} We investigate the effect of our operator fusion introduced in Sec.~\ref{sec_schedule_optim}. One convolutional block in the first stage of a LAS-ResNet-101 ($S$=4, $r$=0.6) is tested. The results in Table~\ref{ablation_op_fusion} validate that every step of operator fusion benefits the practical latency of a block, as the overhead on memory access is effectively reduced. Especially, the fusion of the masker operation and the first convolution is crucial to reducing the latency.

% It is worth noting that increasing the granularity $S$ does not always improve the inference efficiency on other computing devices. For example, $S_{\mathrm{net}}$=4-4-2-1 achieves a sweetspot on Nvidia Jetson TX2 (see \figurename~\ref{fig_resnets} middle).
% More results of ablation studies on other hardware platforms are presented in the supplementary material.

% \begin{wrapfigure}{r}{7cm}
% \centering
% \includegraphics[width=7cm]{ablation_result_resnets}
% \caption{Ablation studies on granularity settings.}
% \label{fig_ablation_sg}
% \end{wrapfigure}
% 

% \noindent\textbf{KD hyperparameters.}
% We perform simple hyperparameter tuning for our self distillation loss (\emph{i.e.} $\beta$ and $T$ in Eq.~\ref{eq_loss}). The tested model is dyn-RegNet-800M, with activation rate target $t=0.7$. From the results in Table~\ref{ablation_KD}, we can observe that self distillation can significant improves the network performance, and $T=4, \beta=0.5$ is superior to other configurations. Therefore, we fix this setting in all our experiments, including our \nameShort~and the DynConv variants \cite{verelst_dynamic_2020}.

\subsection{Visualization}\label{sec_vis}
% \vskip -0.1in
We visualize the masks generated by our masker in the third block of a LAS-ResNet-101 ($S_{\mathrm{net}}$=4-4-2-1) in \figurename~\ref{fig_vis}. The brilliant areas correspond to the locations of 1 elements in a mask, and the computation on the dimmed regions is skipped by our dynamic model. We can witness that the masker accurately locate the most task-related regions (even the tiny aircraft at the corner), which helps reduce the unnecessary computation on background areas. These resultsalso  suggest that for the first stage, the granularity $S$=4 is sufficiently flexible to recognize the important regions, and a \emph{win-win} can be achieved between accuracy and efficiency. Interestingly, the masker could select some objects that are \emph{not labeled} for the sample, \emph{e.g.}, the flower beside the hummingbird and the human holding the camera. This indicates that our spatial-wise dynamic networks can automatically recognize the regions with semantics, and their capability is not limited by the classification labels. This property is helpful in some downstream tasks, such as object detection (Sec.~\ref{sec_det}) and instance segmentation (Sec.~\ref{sec_ins_seg}), which require detecting multiple classes and objects in an image. More visualization results could be found in Appendix~\ref{sec_more_vis}.

\subsection{COCO Object detection}\label{sec_det}
\vskip -0.1in
We further evaluate our \nameShort~on the COCO \cite{lin2014microsoft} object detection task. 
% The COCO dataset \cite{lin2014microsoft} contains 118k training images and 5k images for validation. 
The mean average precision (mAP), the average backbone FLOPs, and the average backbone latency on the validation set are used to measure the network performance. We test two commonly used detection frameworks: Faster R-CNN \cite{ren2015faster} with Feature Pyramid Network \cite{lin2017feature} and RetinaNet \cite{lin2017focal}. Thanks to the generality of our method, we can conveniently replace the backbones with ours pre-trained on ImageNet, and the whole models are finetuned on COCO with the standard setting for 12 epochs (see detailed setup in Appendix~\ref{settings_for_det_seg}). The input images are resized to a short side of 800 and a long side not exceeding 1333. The results of our LAS-ResNet-101 with different $S_{\mathrm{net}}$ settings are presented in Table \ref{tab:coco}. We can observe from the results that when setting the training target as 0.4, the latency of our LAS-ResNet-101 with $S_{\mathrm{net}}$=4-4-7-1 is significantly lower than the static baseline on all devices without sacrificing mAP in both detection frameworks. With a larger training targets, our LASNet can increase the mAP by 0.9\% and 0.8\% in Faster R-CNN and RetinaNet respectively, while still being faster than the baseline method.
% It could be observed that our \nameShort~can reduce the practical latency on GTX1080 and TX2 by 27\% and 45\% respectively while improving the mAP of both detection frameworks.
\begin{table}
%   \vskip -0.2in
  \caption{Object detection results on the COCO dataset.}
%   \vskip -0.1in
  \label{tab:coco}
  \begin{center}
    % \vskip -0.08in
  \resizebox{0.9\linewidth}{!}{
  \begin{tabular}{c c c c c c c}
  % \vskip -0.1in
  \toprule
  Detection & \multirow{2}{*}{Backbone} & Backbone & \multicolumn{3}{c}{Backbone Latency (ms)} & \multirow{2}{*}{mAP (\%)} \\
  \cmidrule{4-6}
  Framework & & FLOPs (G) & V100 & GTX1080 & TX2 & \\
  \midrule
   \multirow{5}{*}{Faster R-CNN} & ResNet-101 (Baseline)  & 141.2 & 39.5 & 119.3 & 729.4 & 39.4 \\ 
   \cmidrule{2-7}
   & \cellcolor{lightgray!50}LAS-ResNet-101 ($S_{\mathrm{net}}$=4-4-2-1, $t$=0.6) & \cellcolor{lightgray!50}90.7  & \cellcolor{lightgray!50}33.8 & \cellcolor{lightgray!50}90.6 & \cellcolor{lightgray!50}524.9 & \cellcolor{lightgray!50}\textbf{40.3} \\
   & \cellcolor{lightgray!50}LAS-ResNet-101 ($S_{\mathrm{net}}$=4-4-2-1, $t$=0.5) & \cellcolor{lightgray!50}79.3  & \cellcolor{lightgray!50}30.7 & \cellcolor{lightgray!50}82.5 & \cellcolor{lightgray!50}477.1 & \cellcolor{lightgray!50}39.8 \\
   & \cellcolor{lightgray!50}LAS-ResNet-101 ($S_{\mathrm{net}}$=4-4-7-1, $t$=0.5) & \cellcolor{lightgray!50}79.5  & \cellcolor{lightgray!50}29.0 & \cellcolor{lightgray!50}79.9 & \cellcolor{lightgray!50}464.8 & \cellcolor{lightgray!50}40.0 \\
   & \cellcolor{lightgray!50}LAS-ResNet-101 ($S_{\mathrm{net}}$=4-4-7-1, $t$=0.4) & \cellcolor{lightgray!50}\textbf{67.9}  & \cellcolor{lightgray!50}\textbf{25.3} & \cellcolor{lightgray!50}\textbf{69.1} & \cellcolor{lightgray!50}\textbf{401.3} & \cellcolor{lightgray!50}39.5 \\
   \midrule
   \multirow{4}{*}{RetinaNet} & ResNet-101 (Baseline)     & 141.2 & 39.5 & 119.3 & 729.4 & 38.5 \\ 
   \cmidrule{2-7}
   & \cellcolor{lightgray!50}LAS-ResNet-101 ($S_{\mathrm{net}}$=4-4-2-1, $t$=0.5) & \cellcolor{lightgray!50}77.8   &\cellcolor{lightgray!50}30.4    &\cellcolor{lightgray!50}81.8  &\cellcolor{lightgray!50}472.6     & \cellcolor{lightgray!50}\textbf{39.3} \\
   & \cellcolor{lightgray!50}LAS-ResNet-101 ($S_{\mathrm{net}}$=4-4-7-1, $t$=0.5) & \cellcolor{lightgray!50}79.4    &\cellcolor{lightgray!50}28.9   &\cellcolor{lightgray!50}79.9 &\cellcolor{lightgray!50}464.8     &\cellcolor{lightgray!50}\textbf{39.3}     \\
   & \cellcolor{lightgray!50}LAS-ResNet-101 ($S_{\mathrm{net}}$=4-4-7-1, $t$=0.4) & \cellcolor{lightgray!50}\textbf{66.4}    &\cellcolor{lightgray!50}\textbf{25.3}   &\cellcolor{lightgray!50}\textbf{69.1} &\cellcolor{lightgray!50}\textbf{401.3}     &\cellcolor{lightgray!50}38.9     \\
  \bottomrule
  \end{tabular}
    }
  \end{center}
  \vskip -0.2in
\end{table}
\begin{table}
%   \vskip -0.2in
  \caption{Instance Segmentation results on the COCO dataset.}
%   \vskip -0.1in
  \label{tab:coco_seg}
  \begin{center}
    % \vskip -0.08in
  \resizebox{\linewidth}{!}{
  \begin{tabular}{c c c c c c c c }
  % \vskip -0.1in
  \toprule
  Segmentation & \multirow{2}{*}{Backbone} & Backbone & \multicolumn{3}{c}{Backbone Latency (ms)} & \multirow{2}{*}{$\mathrm{AP}^\mathrm{mask}$ (\%)} & \multirow{2}{*}{$\mathrm{AP}^\mathrm{box}$ (\%)} \\
  \cmidrule{4-6}
  Framework & & FLOPs (G) & V100 & GTX1080 & TX2 & & \\
  \midrule
  \multirow{5}{*}{Mask R-CNN} & ResNet-101 (Baseline)  & 141.2 & 39.5 & 119.3 & 729.4 & 36.1 & 40.0\\ 
  \cmidrule{2-8}
  & \cellcolor{lightgray!50}LAS-ResNet-101 ($S_{\mathrm{net}}$=4-4-2-1, $t$=0.5) & \cellcolor{lightgray!50}80.5  & \cellcolor{lightgray!50}31.1 & \cellcolor{lightgray!50}83.3 & \cellcolor{lightgray!50}481.9 & \cellcolor{lightgray!50}\cellcolor{lightgray!50}\textbf{37.0} & \cellcolor{lightgray!50}\cellcolor{lightgray!50}\textbf{41.0} \\
  & \cellcolor{lightgray!50}LAS-ResNet-101 ($S_{\mathrm{net}}$=4-4-2-1, $t$=0.4) & \cellcolor{lightgray!50}69.2  & \cellcolor{lightgray!50}27.9 & \cellcolor{lightgray!50}74.8 & \cellcolor{lightgray!50}{431.6} & \cellcolor{lightgray!50}\cellcolor{lightgray!50}36.1 & \cellcolor{lightgray!50}\cellcolor{lightgray!50}40.0 \\
  & \cellcolor{lightgray!50}LAS-ResNet-101 ($S_{\mathrm{net}}$=4-4-7-1, $t$=0.4) & \cellcolor{lightgray!50}\textbf{68.8}  & \cellcolor{lightgray!50}\textbf{25.8} & \cellcolor{lightgray!50}\textbf{70.9} & \cellcolor{lightgray!50}\textbf{411.8} & \cellcolor{lightgray!50}36.2 & \cellcolor{lightgray!50}\cellcolor{lightgray!50}{40.0} \\
%   & \cellcolor{lightgray!50}LAS-ResNet-101 ($S_{\mathrm{net}}$=4-4-7-1, $t$=0.5) & \cellcolor{lightgray!50}82.0  & \cellcolor{lightgray!50}29.5 & \cellcolor{lightgray!50}81.4 & \cellcolor{lightgray!50}474.6 & \cellcolor{lightgray!50}36.6 & \cellcolor{lightgray!50}\cellcolor{lightgray!50}{40.6} \\
  \bottomrule
  \end{tabular}
    }
  \end{center}
  \vskip -0.2in
\end{table}
\subsection{COCO instance segmentation}\label{sec_ins_seg}

We also present the results of instance segmentation on COCO, which demonstrate the effectiveness of our \nameShort~on the dense prediction task. From the results in Table~\ref{tab:coco_seg}, we can observe that when setting the training target as 0.4, the Mask R-CNN \cite{he2017mask} models ($S_{\mathrm{net}}$=4-4-2-1 and $S_{\mathrm{net}}$=4-4-7-1) runs faster on all tested hardware devices without sacrificing the performance. With a training target of 0.5, the $\mathrm{AP}^{\mathrm{mask}}$ and $\mathrm{AP}^{\mathrm{box}}$ of the Mask R-CNN model could be increased by 0.9\% and 1.0\% respectively while still running faster than the baseline method.

%!TEX root = main.tex

\section{Conclusion}\label{sec_conclusion}
\vskip -0.1in
In this paper, we propose to build \emph{latency-aware} spatial-wise dynamic networks (\nameShort) under the guidance of a \emph{latency prediction model}. By simultaneously considering the algorithm, the scheduling strategy and the hardware properties, we can efficiently estimate the practical latency of spatial-wise dynamic operators on arbitrary computing platforms. Based on the empirical analysis on the relationship between the latency and the \emph{granularity} of spatially adaptive inference, we optimize both the algorithm and the scheduling strategies to achieve realistic speedup on many multi-core processors, \emph{e.g.}, the Tesla V100 GPU and the Jetson TX2 GPU. Experiments on image classification, object detection and instance segmentation tasks validate that the proposed method significantly improves the practical efficiency of deep CNNs, and outperforms various competing approaches. 
% The limitation of our work is discussed in the supplementary material.
% The current limitation might be that our latency-ware framework is only constructed for spatial-wise dynamic networks. Support for more types of dynamic models (\emph{e.g.}, channel skipping) will be explored in the future.

% \noindent\textbf{Potential negative social impact.} The optimization of our non-differentiable maskers often requires a proper setting of the temperature. Although we have released a standard setting, the application on other downstream tasks might still require a tuning process, which may result in increased carbon emissions. Future work may additionally explore the possibility of efficient training.

\section*{Acknowledgement}\label{ack}
This work is supported in part by the National Key R\&D Program of China under Grant 2020AAA0105200, the National Natural Science Foundation of China under Grants 62022048 and the Tsinghua University-China Mobile Communications Group Co.,Ltd. Joint Institute..

%%%%%%%%%%%%%%%%%%%%%%%%%%%%%%%%%%%%%%%%%%%%%%%%%%%%%%%%%%%%
\bibliographystyle{plain}
\bibliography{ref}

\clearpage
\appendix

\section*{Appendix}

\section{Latency prediction model.}\label{detailed_latency_predict} 
As the dynamic operators in our method have not been supported by current deep learning libraries, we propose a latency prediction model to efficiently estimate the real latency of these operators on hardware device. The inputs of the latency prediction model include: 1) the structural configuration of a convolutional block, 2) its activation rate $r$ which decides the computation amount, 3) the spatial granularity $S$, and 4) the hardware properties mentioned in Table~\ref{tab_hardware_property}. The latency of a dynamic block is predicted as follows.

% The latency prediction model is 
\textbf{Input/output shape definition}. The first step of predicting the latency of an operation is to calculate the shape of input and output. 
Taking the gather-conv2 operation as an example, the input of this operation is the activation with the shape of $C_{in} \times H \times W $, where $C_{in}$ is the number of input channels, and $H$ and $W$ are the resolution of the feature map.
The shape of the output tensor is  $ P \times C_{out} \times S \times S $, where $P$ is the number of output patches, $C_{out}$ is the number of output channels and $S$ is the spatial granularity. Note that $P$ is obtained based on the output of our maskers.

\textbf{Operation-to-hardware mapping.} Next, we map the operations to hardware.
As is mentioned in the paper, we model a hardware device as multiple processing engines (PEs). 
We assign the computation of each element in the output feature map to a PE. 
Specifically, we consecutively split the output feature map into multiple \emph{tiles}. 
The shape of each tile is $T_P \times T_C \times T_{S1} \times T_{S2}$.
These split tiles are assigned to multiple PEs.
The computation of the elements in each tile is executed in a PE. 
We can configure different shapes of tiles. 
In order to determine the optimal shape of the tile, we make a search space of different tile shapes.
% The best shape minimizes the inference time.
The tile shape has 4 dimensions.
The candidates of each dimension are power-of-2 and do not exceed the corresponding dimension of the feature map.

% To simplify this process, we assume that during tile calculation, when the data is transferred to the memory in the PE, the computing unit in the PE can calculate with the maximum throughput. 
\textbf{Latency estimation.} Then, we evaluate the latency of each tile shape in the search space and select the optimal tile shape with the lowest latency. 
The latency includes the \emph{data movement} latency and the \emph{computation} latency: $\ell=\ell_{\mathrm{data}}+\ell_{\mathrm{computation}}$.

1) \emph{Data movement latency $\ell_\mathrm{data}$.} The estimation of the latency for data movement requires us to model the memory system of a hardware device. We model the memory system of hardware as a three-level architecture \cite{hennessy2011computer}: off-chip memory, on-chip global memory, and local memory in PE.
The input data and weight data are first transferred from the off-chip memory to the on-chip global memory. 
% Parts of the elements in the feature maps are selected and transferred to the on-chip global memory according to the mask. 
% The layout of input feature maps in off-chip memory is CHW (channel-height-width).
% Since all channels in one spatial location are continuously stored in off-chip memory, the data movement of a patch is almost continuous.
We assume the hardware can make full use of the off-chip memory bandwidth to simplify the latency prediction model.

After that, the data used to compute the output tiles is transferred from on-chip global memory to the local memory of each PE. 
% Due to the limited size of the local memory, it is necessary to carry the data from the global memory many times. 
% In order to simplify the model, we assume that most of the data transmission time required by PE can be hidden in PE calculation. 
The latency of data movement to local memory is estimated by its \emph{bandwidth} and \emph{efficiency}. 
We assume each PE only moves the corresponding input feature maps and weights once to compute a output tile so as to simplify the prediction model.
The input data movement latency $\ell_\mathrm{in}$ is calculated by adding the time from off-chip memory to on-chip global memory and the time from on-chip global memory to local-memory together: $\ell_\mathrm{in}=\ell_\mathrm{off2on} + \ell_\mathrm{global2local}$.
Contrary to the input data, the output data $\ell_\mathrm{out}$ are moved from local memory to on-chip global memory and then to off-chip memory: $\ell_\mathrm{out}=\ell_\mathrm{local2global} + \ell_\mathrm{on2off}$.
We calculate the total data movement latency by adding the input and output data movement latency together: $\ell_\mathrm{data}=\ell_\mathrm{in} + \ell_\mathrm{out}$.

The latency of data movement is affected by the granularity $S$: when the granularity $S$ is small, the same input data has a higher probability of being sent to multiple PEs to compute different output patches, which significantly increases the number of on-chip memory movement.
And due to the small amount of data transmitted each time and the data is randomly distributed, the efficiency of data movement will be low. 
This accounts for our experiment results in the paper that \emph{a larger $S$ will effectively improve the practical efficiency}.

2) \emph{Computation latency $\ell_\mathrm{computation}$.} The computation latency of each tile is estimated using the PE's \emph{maximum throughput of FP32 computation} and the \emph{FLOPs} of computing an output tile. The total computation latency can be obtained according to the number of tiles and the number of PEs. 

To summarize, our latency prediction model can predict the real latency of dynamic operators by considering both the \emph{data movement} cost and the \emph{computation} cost.
Guided by the latency prediction model, we propose our LASNets with coarse-grained spatially adaptive inference ($S>1$).  It is validated in our paper that LASNets achieve better efficiency than previous approaches \cite{xie2020spatially, verelst_dynamic_2020} ($S=1$), as it effectively reduces the data movement latency, which is rarely considered by other researchers.

\section{Detailed experimental settings}\label{sec_detailed_settings}
In this section, we present the detailed experiment settings which are not provided in the main paper due to the page limit.
\subsection{Latency prediction}\label{sec_detailed_settings_for_latency_predict}

\noindent\textbf{Hardware properties} considered by our latency prediction model include the number of processing engines (\#PE), the floating-point computation in a processing engine (\#FP32), the frequency and the bandwidth. We test four types of hardware devices, and their properties are listed in Table~\ref{tab_hardware_property}.

\begin{table}[h]
  \caption{Hardware properties.}
%   \vskip -0.1in
  \label{tab_hardware_property}
  \centering
  \begin{tabular}{ccccc}
    \toprule
    
    Name & \#PE     &    \#FP32  &   frequency (MHz)   & bandwidth (G) \\
    \midrule
    Nvidia Tesla V100    & 80 & 64  & 1500  & 700  \\
    Nvidia GTX1080       & 20 & 64  & 1700  & 320  \\
    Nvidia Jetson TX2    & 2  & 128 & 1300  & 59.7 \\
    Nvidia Nano          & 1  & 128 & 921   & 25.6 \\
    \bottomrule
  \end{tabular}
% \vskip -0.1in
\end{table}

It could be found that the server GPU V100 is the most powerful hardware device, especially with the most number of processing engines (\#PE). Therefore, spatially adaptive inference could easily fall into a \emph{memory-bounded} operation on V100 due to its high parallelism. Our experiment results in \figurename~7 (a) and \figurename~8 in the paper can reflect this phenomenon: the more flexibility the computation is, the harder to improve the practical efficiency. 

In contrast, on the less powerful computing devices such as the IoT devices, the real acceleration is close to the theoretical effect (compare \figurename~7 (a) left with \figurename~7 (a) middle).

\noindent\textbf{Operator fusion.} 

1) \emph{Fusing the masker and the first convolution.} We mentioned in Sec.~3.4 of the paper that the masker operation is fused with the first 1$\times$1 convolution in a block to reduce the cost on memory access. This is feasible because the two operators share the same input feature, and their convolutional kernel sizes are both 1$\times$1. 

Note that during the inference stage, we only need to perform $\arg\max$ along the channel dimension of a mask $\mathbf{M}\in\mathbb{R}^{2\times H\times W}$ to obtain the positions of the gathered pixels. Therefore, we can reduce the output channel number of our maskers from 2 to 1 since the convolution is a linear operation: 
\begin{equation}
 [\mathbf{x}*\mathbf{W}]_{:,:,0}>[\mathbf{x}*\mathbf{W}]_{:,:,1} \Longleftrightarrow \mathbf{x}*(\mathbf{W}_{:,:,0}-\mathbf{W}_{:,:,1}) > 0.
\end{equation}

Afterwards, we fuse the masker with the first convolution layer by performing once convolution whose output channel number is $C+1$, where $C$ is the original output width of the first convolution. The output of this step is split into a feature map (for further computation) and a mask (for obtaining the index for gathering). Such operator fusion avoids the repeated reading the input feature, and helps reduce the inference latency (see Table 1 in the paper).

2) \emph{Fusing the gather operation and the dynamic convolution.}
To facilitate the scheduling on hardware devices with multiple PEs, the masker generates the indices of activated patches instead of sparse mask at inference time. 
In this way, it is easy to evenly distribute the computation of output patches to different PEs, thus avoiding unbalanced computation of PEs.
Each element in the indices represents the index of an activated patch. 
PE fetches the input data from the corresponding positions on the feature map according to the index.
The output patches could be densely stored in memory. Such operator fusion benefits the contiguous memory access and parallel computation on multiple PEs.

3) \emph{Fusing the scatter operation and the add operation.}

Similar to the previous operation, each PE fetches a tile of data from the residual feature map according to the index, adds them with the corresponding feature map from previous dynamic convolution, and then stores the results to the corresponding position on the residual feature map according to the index.
This optimization can significantly reduce the costs on memory access.

\begin{figure}
% \vskip -0.2in
     \centering
     \begin{subfigure}[b]{\textwidth}
         \centering
         \includegraphics[width=\textwidth]{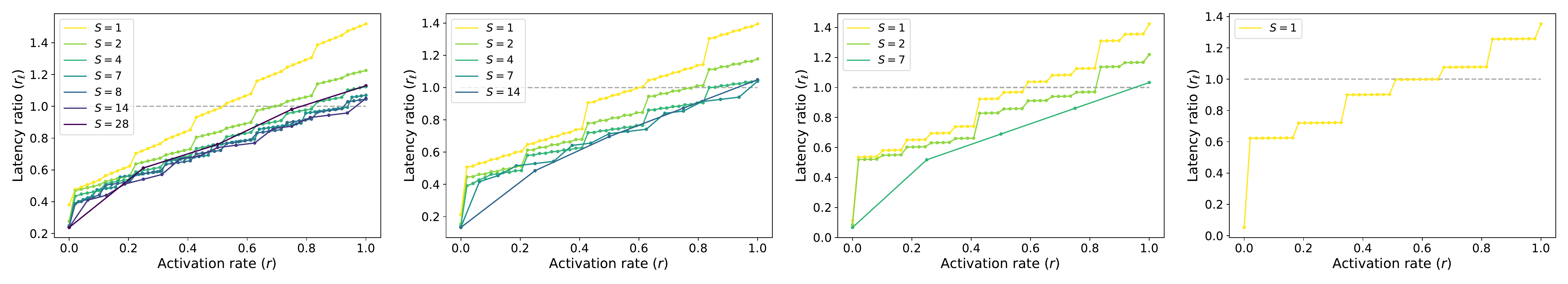}
        %  \vskip -0.1in
         \caption{Relationship between $r_{\ell}$ and $r$ for LAS-ResNet blocks on Nvidia GeForce GTX1080.}
         \label{fig_r_r_1080}
     \end{subfigure}
     \begin{subfigure}[b]{\textwidth}
         \centering
         \includegraphics[width=\textwidth]{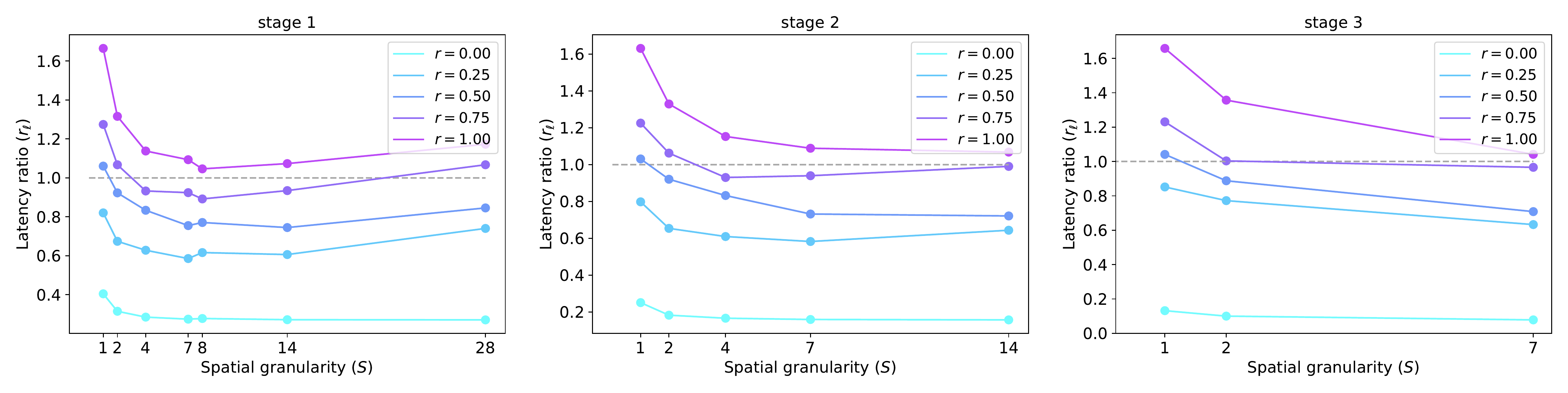}
        %  \vskip -0.1in
         \caption{Relationship between $r_{\ell}$ and $S$ for LAS-ResNet blocks on Nvidia GeForce GTX1080.}
         \label{fig_r_S_1080}
     \end{subfigure}
     \caption{Latency prediction results of LAS-ResNet blocks on the Nvidia GTX1080 GPU.}
        \label{fig_latency_r_s}
\end{figure}

\noindent\textbf{Speed test.} We test the latency on real hardware devices to evaluate the accuracy of our latency prediction model.
On GPUs, we use TVM \cite{chen2018tvm} and CUDA (version 11.6) for code generation and compilation respectively. The results in Fig. 4 of the paper validate the effectiveness of our model.

\subsection{ImageNet classification}
As mentioned in the paper, we use pre-trained CNN models in the official torchvision website to initialize our backbone parameters, and finetune the overall models for 100 epochs. The initial learning rate is set as 0.01$\times$batch size/128, and decays with a cosine shape. The training batch size is determined on the model size and the GPU memory. For example, we train our LAS-ResNet-101 on 8 RTX 3090 GPUs with the batch size of 512, and the batch size for LAS-ResNet-50 is doubled. We use the same weight decay and the standard data augmentation as in the RegNet paper \cite{radosavovic2020designing}. For our own hyper-parameter $\tau$ in Eq.~(1) of the paper, this Gumbel temperature $\tau$ exponentially decreases from 5 to 0.1 in the training procedure. For the training hyper-parameter in Eq. (2), we simply fix $\alpha=10,\beta=0.5$ and $T=4.0$ for all dynamic models. We conduct a very simple grid search with a RegNet for $\beta\in\{0.3,0.5\}$ and $T\in\{1.0,4.0\}$ to determine their values.

\begin{figure}
% \vskip -0.2in
     \centering
     \begin{subfigure}[b]{0.6\textwidth}
         \centering
         \includegraphics[width=\textwidth]{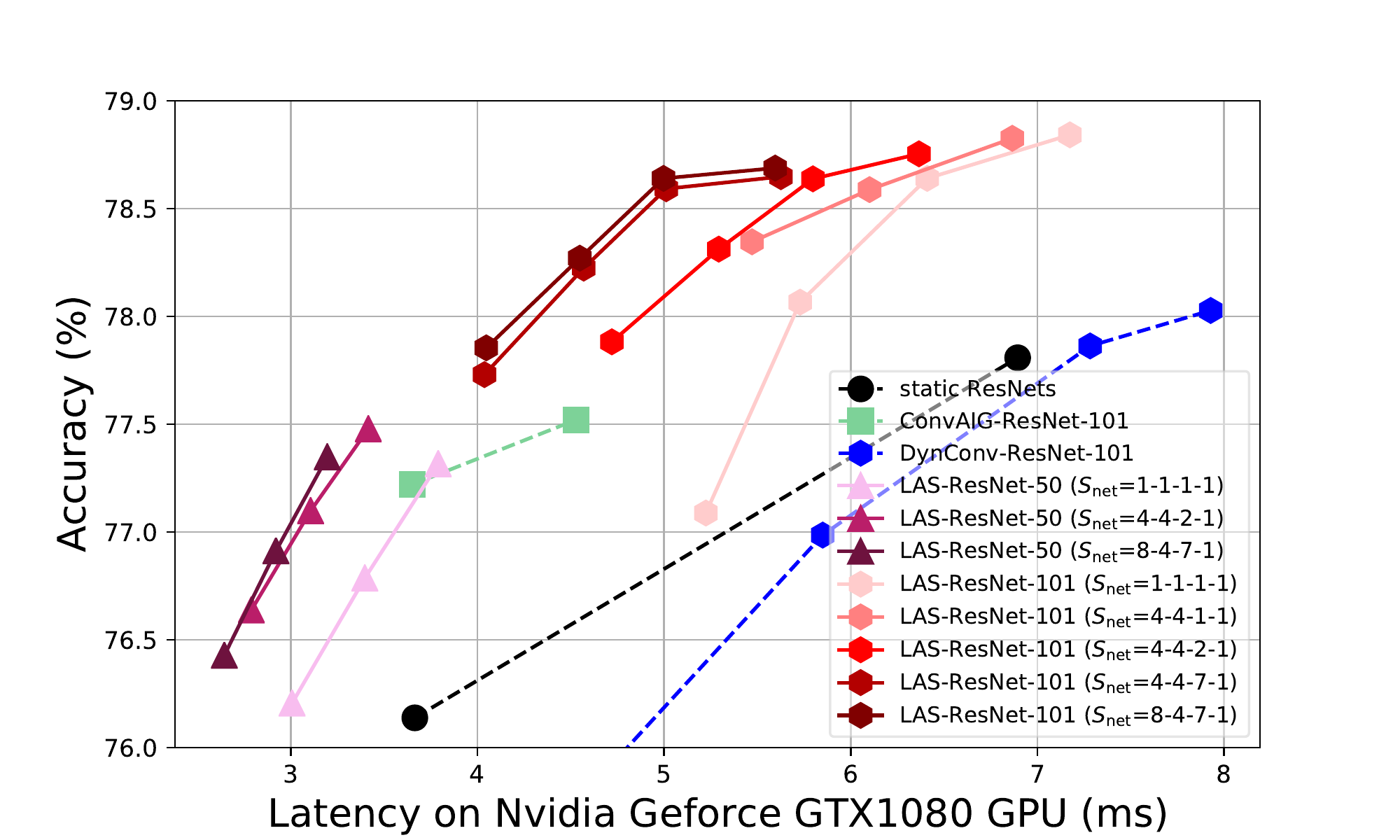}
        %  \vskip -0.1in
         \label{fig_r_r_1080}
     \end{subfigure}
     \caption{Experimental results on the ImageNet classification task.}
        \label{fig_latency_acc_1080}
        % \vskip -0.1in
\end{figure}
\begin{figure}
     \centering
    %  \vskip -0.2in
         \centering
         \includegraphics[width=\textwidth]{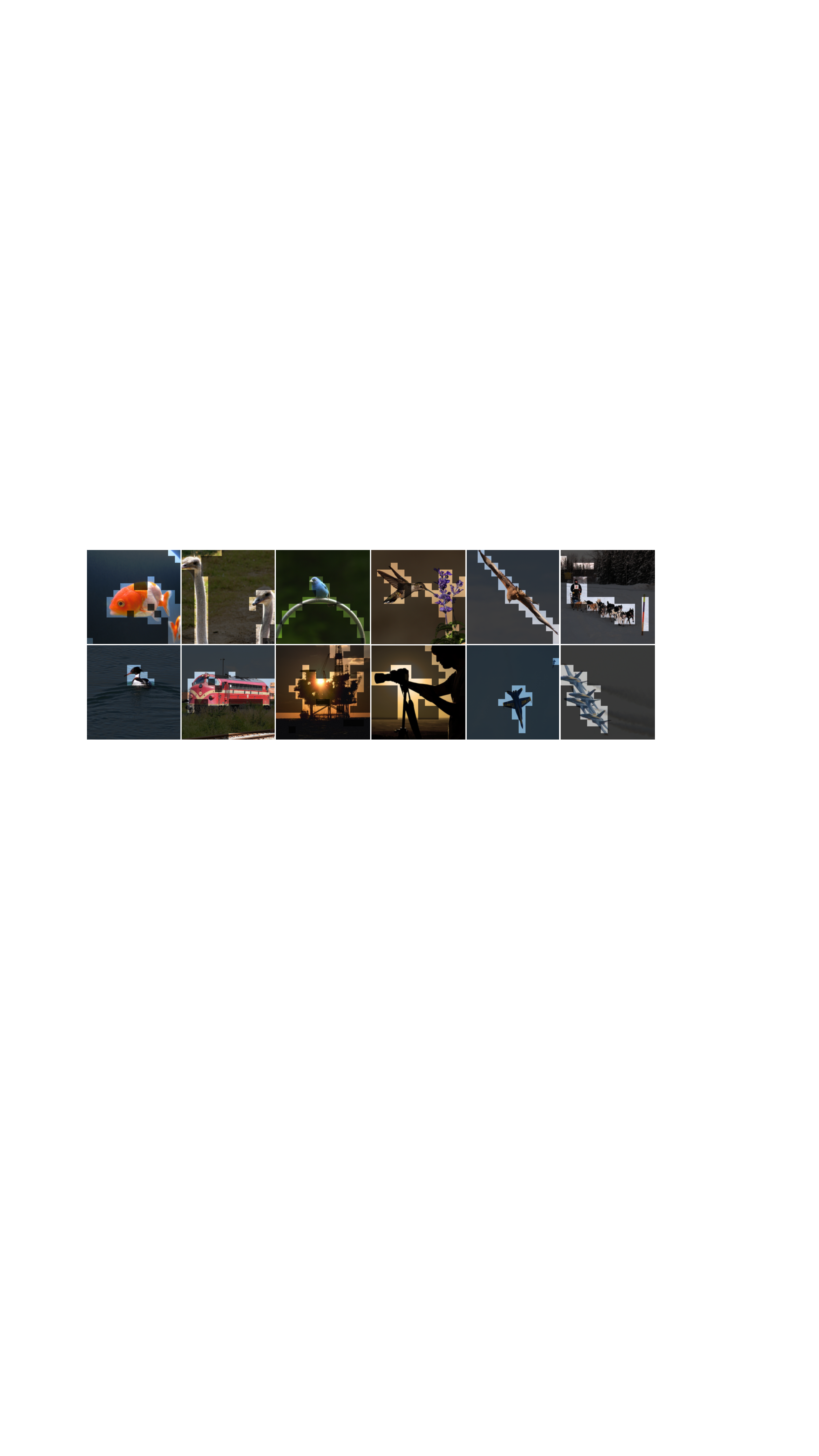}
        %  \vskip -0.05in
         \caption{Visualization results.}
         \vskip -0.1in
         \label{fig_vis}
\end{figure}

\subsection{COCO object detection \& instance segmentation } \label{settings_for_det_seg}
We use the standard setting suggested in \cite{lin2017feature,lin2017focal,he2017mask}, except that we decrease the learning rate for our pre-trained backbone network. We simply set a learning rate multiplier 0.5 for Faster R-CNN \cite{ren2015faster}, 0.2 for RetinaNet \cite{lin2017focal} and 0.5 for Mask R-CNN \cite{he2017mask}. As for the additional loss items, the hyper-parameters are kept the same as training our classification models, except that the temperature is fixed as 0.1 in the 12 training epochs.

\section{More experimental results}
In this section, we report more experimental results which are not presented in the main paper.
\subsection{Latency prediction}\label{supp_results_latency_pred}
In \figurename~5 and \figurename~6 of the paper, we report the latency prediction results of LAS-ResNet on V100 and TX2. Here we present the results on GTX1080 (\figurename~\ref{fig_latency_r_s}). It can be found that $S_\mathrm{net}$=8-4-7-1 (which is the same to the optimal setting on V100 in the paper) will lead to faster inference on GTX1080. This is reasonable since GTX1080 has a large \#PE than those IoT devices, and requires more contiguous memory access (a larger $S$ for coarse-grained spatially adaptive inference) to achieve realistic speedup ($r_{\ell}<1$). The accuracy-latency curves in \figurename~\ref{fig_latency_acc_1080} further validate this observation.

\subsection{ImageNet classification}\label{supp_results_IN_cls}
\textbf{Results on GTX1080 GPU.} In \figurename~\ref{fig_main_results} of the paper, we report the ImageNet classification results of LAS-ResNets on V100 and TX2.
Here we present the results of LAS-ResNet on GTX1080 (\figurename~\ref{fig_latency_acc_1080}). From the results in \figurename~\ref{fig_latency_acc_1080}, we can get the same conclusion as in the paper, that a proper large $S$ is more parallel-friendly than the finest granularity. Remarkably, the latency of ResNet-101
could be reduced by 41\% without sacrificing the accuracy on GTX1080 under the spatial granularity setting of $S_{\mathrm{net}}$=8-4-7-1. With a similar inference latency, the accuracy of a static ResNet-101 could be increased by 1.0\% by our LAS-ResNet-101 ($S_{\mathrm{net}}$=4-4-2-1).

\textbf{Results of the extreme situation of} $S_{\mathrm{net}}$=56-28-14-7. We mentioned in the paper that when we set $S_{\mathrm{net}}$=56-28-14-7, the spatially adaptive inference paradigm will reduce into layer skipping. We experiment on ResNet-50, and find that although being slightly faster (13ms) than our LAS-ResNet-50 ($S_{\mathrm{net}}$=4-4-2-1, 16ms), the accuracy of LAS-ResNet-50 with $S_{\mathrm{net}}$=56-28-14-7 could be significantly degraded from 76.6\% (ours, $S_{\mathrm{net}}$=4-4-2-1) to 76.1\%. Therefore, we mainly focus on the discussion of spatially adaptive inference in this paper.

\subsection{More visualization results}\label{sec_more_vis}
In addition to \figurename~9 in the paper, here we present more visualization results of the regions selected by our masker in the 3-rd block of a LAS-ResNet-101 ($S_{\mathrm{net}}$=4-4-2-1) in \figurename~\ref{fig_vis}, which demonstrate that our spatially adaptive inference paradigm can effectively locate the most task-related areas in image features, and reduce the unnecessary computation on those background areas.

\section{Limitations}\label{sec_limitation}
The current limitations of our work include the following aspects:

1) the latency-ware co-designing framework is only constructed for spatial-wise dynamic networks. Support for more types of dynamic models (\emph{e.g.} channel skipping) will be explored in the future;

2) To achieve faster inference, the spatial masks are defined the same for all input/output channels, which might limit the flexibility of adaptive inference. Future work may explore more flexible forms of dynamic computation;

3) The combination with other acceleration techniques such as Winograd, and the implementation on more CNN backbones may be worth studying in the future. 

\noindent\textbf{Social impact.} Our work can help reduce the inference cost of deep CNNs, but the training of our models might potentially increase the carbon emissions.

\end{document}